\documentclass[10pt,twocolumn,letterpaper]{article}

\usepackage{iccv}
\usepackage{times}
\usepackage{epsfig}
\usepackage{graphicx}
\usepackage{amsmath}
\usepackage{amssymb}

\usepackage{epsfig}
\usepackage{graphicx}
\usepackage[linesnumbered,ruled,vlined,lined,boxed,commentsnumbered]{algorithm2e}
\usepackage{caption}
\usepackage{indentfirst}
\usepackage{multirow}
\usepackage{booktabs}
\usepackage{threeparttable}
\usepackage{boldline}
\usepackage{float}
\usepackage{makecell}
\usepackage{caption}
\newcommand{\udots}{\mathinner{\mskip1mu\raise1pt\vbox{\kern7pt\hbox{.}}\mskip2mu\raise4pt\hbox{.}\mskip2mu\raise7pt\hbox{.}\mskip1mu}}

\usepackage[breaklinks=true,bookmarks=false]{hyperref}

\iccvfinalcopy % *** Uncomment this line for the final submission

\def\iccvPaperID{****} % *** Enter the ICCV Paper ID here

\ificcvfinal\fi

\begin{document}

%%%%%%%%% TITLE
\title{Perspective-Guided Convolution Networks for Crowd Counting}

{
\author{Zhaoyi Yan$^{1}$\footnotemark[2], Yuchen Yuan$^{2}$, Wangmeng Zuo$^{1,3}$\footnotemark[1], Xiao Tan$^{2}$, Yezhen Wang$^{1}$, Shilei Wen$^{2}$, Errui Ding$^{2}$\\
$^1${\small Harbin Institute of Technology}, $^2${\small Department of Computer Vision Technology (VIS), Baidu Inc.}, $^3${\small Peng Cheng Laboratory, Shenzhen} \\
{\tt \small yanzhaoyi@outlook.com, }  {\tt \small wmzuo@hit.edu.cn} \\
{\tt \small \{tanxchong, yezhen.wang0305\}@gmail.com} \\
{\tt \small \{yuanyuchen02, wenshilei, dingerrui\}@baidu.com} \\
}

\maketitle

\renewcommand{\thefootnote}{\fnsymbol{footnote}}
\footnotetext[2]{{This work was done when Zhaoyi Yan was a research intern at Baidu}}
\footnotetext[1]{Corresponding author}

\ificcvfinal\thispagestyle{empty}\fi

%%%%%%%%% ABSTRACT
\begin{abstract}
In this paper, we propose a novel perspective-guided convolution (PGC) for convolutional neural network (CNN) based crowd counting (i.e. PGCNet), which aims to overcome the dramatic intra-scene scale variations of people due to the perspective effect.
While most state-of-the-arts adopt multi-scale or multi-column architectures to address such issue, they generally fail in modeling continuous scale variations since only discrete representative scales are considered.
PGCNet, on the other hand, utilizes perspective information to guide the spatially variant smoothing of feature maps before feeding them to the successive convolutions.
An effective perspective estimation branch is also introduced to PGCNet, which can be trained in either supervised setting or weakly-supervised setting when the branch has been pre-trained.
Our PGCNet is single-column with moderate increase in computation, and extensive experimental results on four benchmark datasets show the improvements of our method against the state-of-the-arts.
Additionally, we also introduce Crowd Surveillance, a large scale dataset for crowd counting that contains 13,000+ high-resolution images with challenging scenarios.
%
%Code and pre-trained models are available at \url{https://github.com/Zhaoyi-Yan/PGCNet}.
\end{abstract}

%###################
% maybe we should add a chart of speed comparison, since we've mentioned computational burden in the abstract
%###################

%%%%%%%%% BODY TEXT
%%%%%%%%% INTRODUCTION
\section{Introduction}
\label{sec:introduction}
The growth of global population and urbanization has been consistently promoting the frequency of crowd gathering.
In such scenarios, stampedes and crushes can be life threatening and should always be prevented.
Congested scene analysis and understanding is thus essential to the management, control, and security guarding of crowd gathering in cities.
Among the developments in congested scene analysis, crowd counting~\cite{zhan2008crowd,li2015crowded} is one of the fundamental tasks, and recently has drawn considerable attention from the computer vision community.

Single image based crowd counting remains an active but challenging topic due to the complex distribution of people, non-uniform illumination, inter- and intra-scene scale variations, cluttering and occlusions, etc.
Existing crowd counting methods can be broadly classified into three categories, \ie detection-based~\cite{ren2015faster,liu2016ssd}, regression-based~\cite{idrees2013multi,wang2015deep}, and CNN-based methods~\cite{li2018csrnet,liu2018leveraging}.
Among them, CNN-based methods have been studied in depth in the past few years, and have achieved superior performances in terms of accuracy and robustness.

\begin{figure}[t]
\centering
\includegraphics[width=0.45\textwidth]{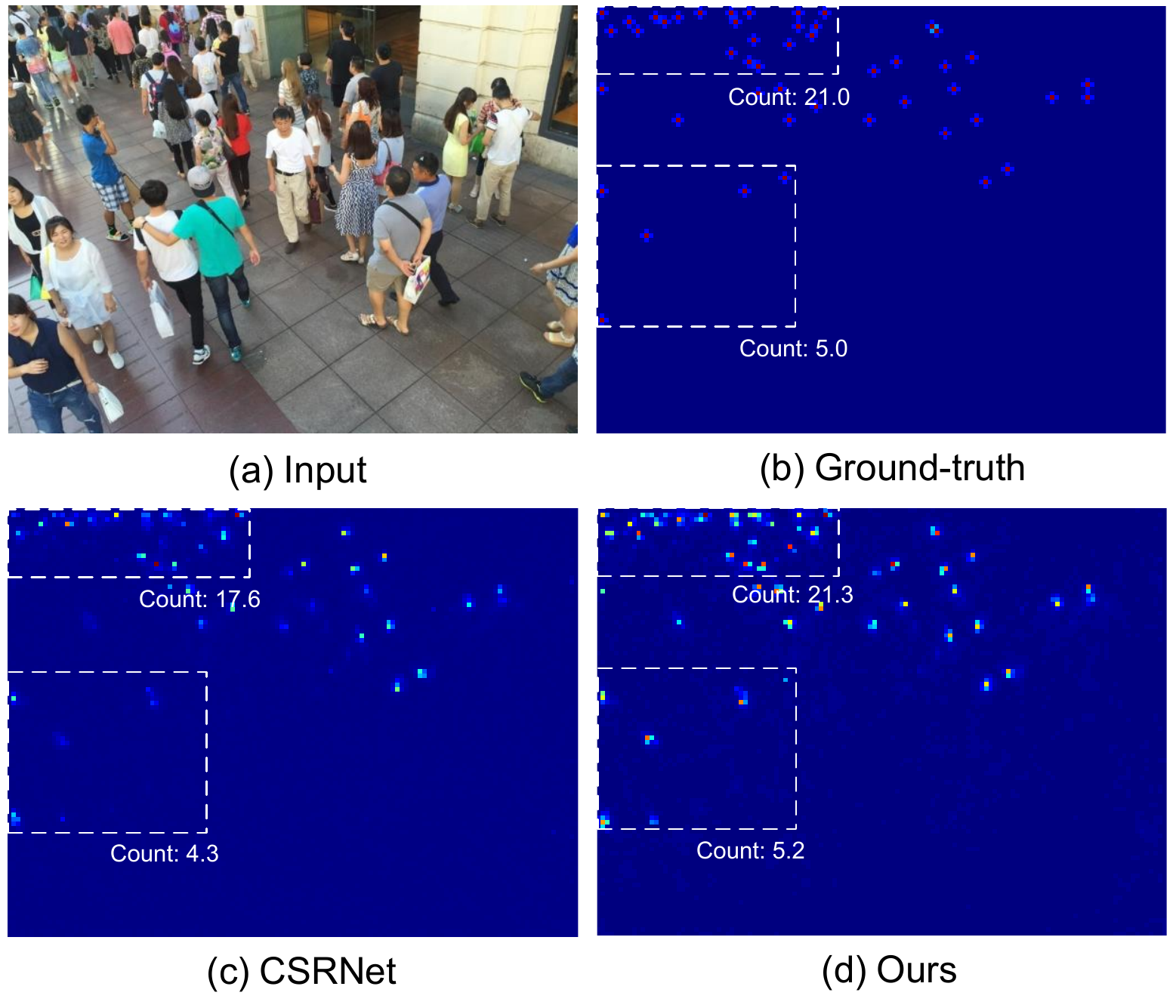}
 \scriptsize
\caption{Density map estimations by CSRNet~\cite{li2018csrnet} and our PGCNet. The MAE of our PGCNet is 2.8, much lower than that of CSRNet (7.7). It is observed that PGCNet has consistent better performances at either smaller or larger scales among the marked regions.}
\label{fig:comparion_with_csrnet}
\vspace{-1em}
\end{figure}

However, the dramatic intra-scene scale variations of people due to the perspective effect forms a major challenge. % the influence of perspective %
Existing methods~\cite{huang2018body, onoro2016towards, sam2017switching, sindagi2017generating,zhang2015cross, zhang2016single} usually adopt multi-scale
or multi-column architectures to fuse the features of different scales.
Yet, they suffer from several limitations.
Firstly, the multi-column architectures (\eg MCNN~\cite{zhang2016single}) are ineffective to train.
As shown in~\cite{li2018csrnet}, MCNN cannot even compete with a deeper CNN due to the high correlation of the features learned by different columns.
Secondly, they only consider discrete scales, which is limited when addressing continuous scale variations in practical scenarios.
Thirdly, the computational cost increases linearly with the growth of columns or scales.

In~\cite{li2018csrnet}, a deeper CNN (CSRNet) with dilated convolutions has achieved state-of-the-art performance.
Nevertheless, it still delivers fixed receptive field for different scales of people, thereby remaining vulnerable to the highly variant intra-scene scales.
It is seen in Fig.~\ref{fig:comparion_with_csrnet}(c) that CSRNet performs well at intermediate scales, but behaves relatively poor at smaller or larger scales.
We thus take a step forward to propose the perspective-guided convolutional network (PGCNet), which is a single-column CNN that aims to tackle the continuous scale variation issue with perspective information considered.

%
% In PGCNet, perspective information is referred as the representation of human scales in a given image, which is used to allocate scale-adaptive receptive fields in the density map % estimation via a single column CNN model.
The perspective information encodes the distance between camera and a scene, which serves as a reasonable scale estimation of people.
We thus adopt it to allocate spatially variant receptive fields, thereby conducting scale adaptive density map estimation.
To this end, we propose a novel perspective-guided convolution (PGC), in which the perspective information functions to guide the spatially variant smoothing of feature maps before taking them to the successive convolutions.
%
%
% Following such idea, local smoothing with larger Gaussian kernel is adopted for people at larger scale, while smaller Gaussian kernel is deployed for people at smaller scale.
As a result, larger (or smaller) Gaussian kernels for feature smoothing are adopted for people at larger (or smaller) scales.
After such spatially variant feature smoothing, the conventional spatially invariant convolution is appended, which forms a PGC block.
It is worth noting that PGC serves as an insertable module to existing architectures, and our PGCNet is formulated by stacking multiple PGC blocks upon a CNN backbone.

However, off-the-rack perspective annotations are seldom available for existing datasets.
We hence introduce a perspective estimation branch to PGCNet, which can be learned either in supervised or weakly-supervised setting when the branch has been pre-trained.

Experimental results on benchmark datasets against the state-of-the-arts show the favorable performance of our proposed PGCNet in handling intra-scene scale variations.
In addition, we also introduce Crowd Surveillance, a large scale dataset for crowd counting that contains 10,000+ high-resolution images with complicated backgrounds and varied crowd counts.
This dataset will be released as a new benchmark to facilitate crowd counting researches.
To sum up, the main contribution of this work includes:

(1) The PGC, as an insertable module, is proposed to handle the intra-scene scale variations of crowd counting;

(2) A perspective estimation branch is introduced, which can be trained with or without perspective annotations;

(3) An end-to-end trainable PGCNet is formulated with (1) and (2);

(4) A new large scale dataset is introduced;

(5) State-of-the-art performance is achieved by PGCNet on four benchmark datasets, \eg 57.0 MAE on ShanghaiTech Part A and 8.8 MAE on ShanghaiTech Part B.

%%%%%%%%%%%%%%%%%%%%%%%%%%%%%%%%%%%%%%%%%%%%%
% Related Work
%
%%%%%%%%%%%%%%%%%%%%%%%%%%%%%%%%%%%%%%%%%%%%%
\section{Related Work}
\label{sec:related_work}
Crowd counting methods can be roughly categorized into three subsets, \ie detection-based, regression-based and CNN-based methods.
In this paper, we only review CNN-based methods, which are the most related to our method.
Besides, the exploration on perspective normalization in crowd counting is also surveyed.

\subsection{CNN-based Methods}
\label{sec:counting_by_cnn}
Benefited from the great success of CNN, many CNN-based works of crowd counting have been proposed in recent years. These methods usually focus on typical techniques, including multi-scale~\cite{zhang2016single, sam2017switching, sindagi2017generating, zhang17sacnn, shi18pacnn, shen2018crowd}, context~\cite{sindagi2017generating}, multi-task~\cite{zhang2015cross, huang2018body, liu2018leveraging}, and others~\cite{liu2018decidenet, liu2018crowd, ranjan2018iterative}.
Recently, more methods have been proposed for handling the scale variation issue.
For instance, Zhang~\etal~\cite{zhang2016single} suggest a multi-column architecture (MCNN) that combines features with different sizes of receptive fields.
In Switching-CNN~\cite{sam2017switching}, one of the three regressors is assigned for an input image in refer to its specific crowd density.
CP-CNN~\cite{sindagi2017generating} incorporates MCNN with local and global contexts.
SANet~\cite{Cao2018Scale} employs scale aggregation modules for multi-scale representation.
And instead of multi-column architecture, CSRNet~\cite{li2018csrnet} enlarges receptive fields by stacking dilated convolutions.

Our proposed method differs from existing CNN-based method in two aspects: firstly, our method is able to handle continuous scale variations of each single pedestrian in the image, instead of simply fusing the features of different scales; secondly, the perspective information is taken into account as a vital estimator of pedestrian scales.

\subsection{Perspective Normalization}
Perspective is originally adopted in the normalization of extracted features from foreground objects in~\cite{chan2008privacy}.
Later, Lempitsky~\etal~\cite{lempitsky2010learning} attempt to deal with perspective distortion by optimizing the loss on the MESA-distance.
Among the CNN-based methods, the perspective information is usually exploited as part of the pre-processing to generate the density maps~\cite{zhang2015cross, zhang2016single, sam2017switching, huang2018body}, but is seldom directly encoded into the network architecture.
One of the most relevant work to our method is PACNN~\cite{shi18pacnn}.
However, PACNN is still based on the multi-column architecture with discrete scales.
In PACNN, two density maps are estimated from two columns based on VGG-16~\cite{simonyan2014very} backbone.
The two predicted density maps are assigned weights generated by the perspective map and later are combined together as the final estimation.
However, we argue that a better choice would be an explicit perspective normalization during the training of the network itself, which is able to tackle the continuous scale variation issue, as illustrated in Sec.~\ref{sec:persp_conv}.

%%%%%%%%%%%%%%%%%%%%%%%%%%%%%%%%%%%%%%%%%%%
%%%          Fig: Net architecture.
%%%%%%%%%%%%%%%%%%%%%%%%%%%%%%%%%%%%%%%%%%%
\begin{figure*}[t]
\begin{center}
\includegraphics[scale=.09]{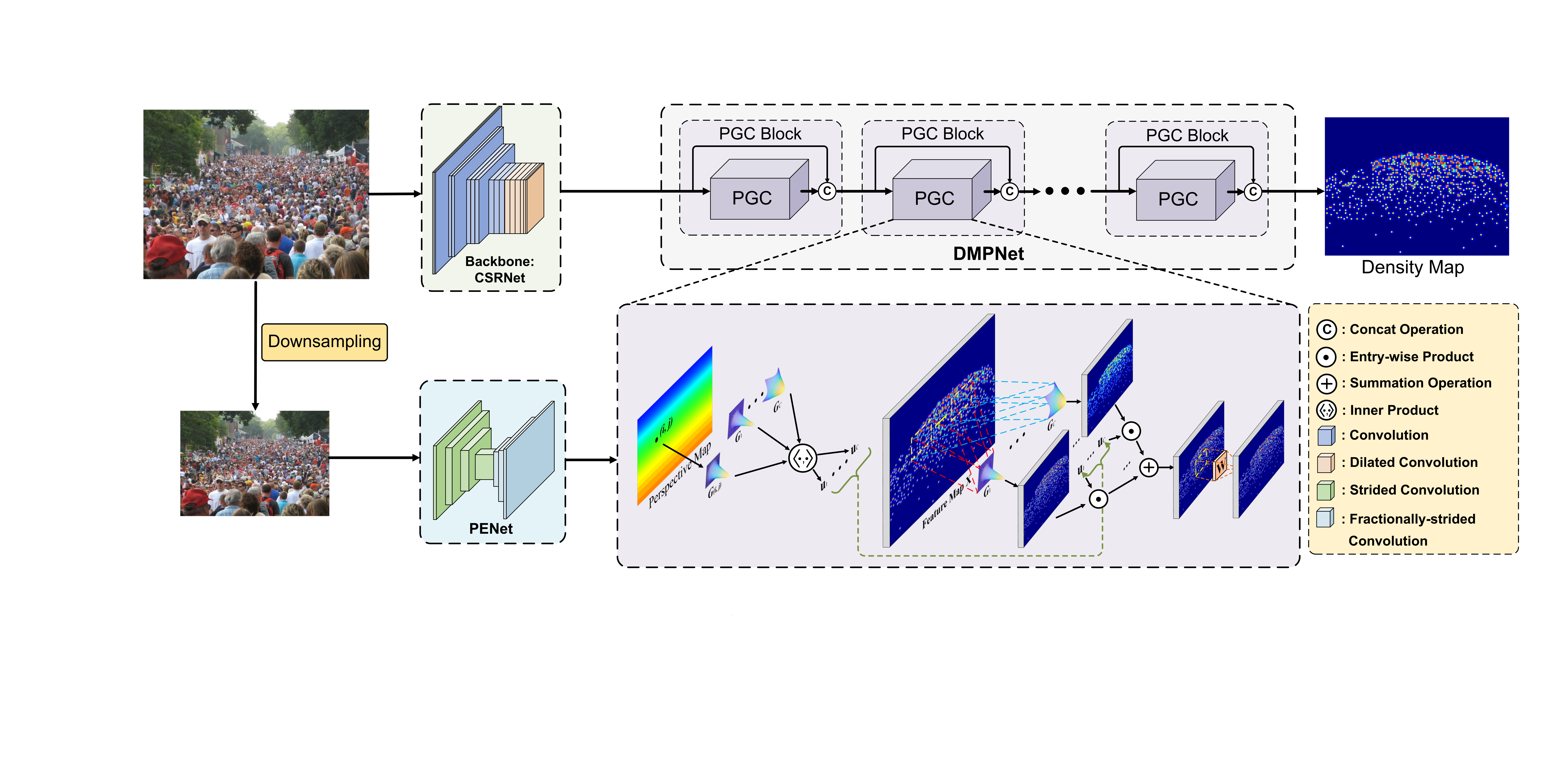}
\end{center}
   \scriptsize
   \vspace{-2.5em}
   \caption{The architecture of the proposed PGCNet. The
                 DMPNet takes the backbone features and the estimated perspective map from PENet to generate the final density map estimation.}
   \label{fig:PGCNet}
\end{figure*}

%%%%%%%%%%%%%%%%%%%%%%%%%%%%%%%%%%%%%%%%%%%%%
% Proposed method
%
%%%%%%%%%%%%%%%%%%%%%%%%%%%%%%%%%%%%%%%%%%%%%
\section{Proposed Method}
\label{sec:proposed_method}
In this section, we first describe the principles of PGC, then introduce a perspective estimation branch for perspective map estimation, and finally provide the network architecture and learning objective of the complete PGCNet.

\subsection{Perspective-Guided Convolution}
\label{sec:persp_conv}
To handle the intra-scene scale variations mentioned in Sec.~\ref{sec:introduction}, it is desirable to use a larger (or smaller) receptive field for people at larger (or smaller) scales.
Let $\mathbf{x}$ and $\mathbf{p}$ be the feature map and perspective map of an image, respectively.
%and
Without loss of generality, $\mathbf{p}$ is downsampled to the same size $h \times w$ with $\mathbf{x}$.
A straightforward way is to apply spatially variant convolutions by assigning scale-aware kernel sizes based on the corresponding perspective values, \ie kernels with different sizes may be used at different locations of a feature map.
However, several drawbacks hinder such solution:
(1) the discrete kernel sizes might be incompatible with the continuous perspective values,
(2) arbitrary spatially variant convolution is hard to implement and optimize, and
(3) additional constraints are required to effectively enforce the consistency among different scales~\cite{shen2018crowd}.

To address the issues above, we propose perspective-guided convolution (PGC), which functions in a two-stage scheme: spatially variant Gaussian filtering (which smooths the feature maps in a spatially variant way) and spatially invariant convolution (\ie the conventional convolution).
To begin with, the perspective map $\mathbf{p}$ is normalized as,
\begin{equation}\label{eqn:ada_sig}
\widetilde{\mathbf{p}} = \zeta\left(\mathbf{p}\right)=\frac{1}{1+e^{-\alpha\left(\mathbf{p}-\beta\right)}}
\end{equation}
where $\zeta\left(x\right)$ is a sigmoid-like function, and $\alpha$ and $\beta$ are two parameters learned during training.
We then define the blur map as,
\begin{equation}\label{eqn:sigma_map}
  {\boldsymbol{\sigma}} = \max(a (\widetilde{\mathbf{p}} - p_0), 0),
\end{equation}
where $a$ and $p_0$ are another two trainable parameters.
Thus, in the spatially variant Gaussian filtering, the smoothing result $\widetilde{x}_{i,j}$ at $(i,j)$ can be obtained by
\begin{equation}\label{eqn:smoothing}
\widetilde{x}_{i,j} = \sum_k \sum_l x_{k,l} G_{\sigma_{i,j}} (i,j,k,l),
\end{equation}
where $G_{\sigma_{i,j}} (i,j,k,l)$ is a Gaussian kernel with standard deviation $\sigma_{i,j}$ centered at $(i,j)$,
\begin{equation}\label{eqn:gaussian}
\small
G_{\sigma_{i,j}} (i,j,k,l) \!=\! \frac{1}{\sqrt{2 \pi }\sigma_{i,j} } \exp \left( - \frac{\left( (k \!-\! i)^2 \!+\! (l \!-\! j)^2 \right)}{2\sigma_{i,j}^2}  \right).
\end{equation}
The perspective-guided convolution can then be defined as,
\begin{equation}\label{eqn:persp_conv}
\mathbf{y} = \mathbf{W}^{T} \widetilde{\mathbf{x}},
\end{equation}
where $\mathbf{W}$ denotes the spatially invariant convolution kernel.

However, Eqn.~(\ref{eqn:smoothing}) is computationally heavy. We thus present an efficient version for approximation.
First, we sample $N$ candidate Gaussian filters with size $K \times K$ and standard deviation in the pre-defined range $[c, d]$.
After that, principal component analysis (PCA) is performed on the candidates to obtain the eigenvectors $\{ G_q\}_{q=1}^C$ corresponding to the $C$ non-zero eigenvalues.
For each $G_q$, we define the coefficient map $\mathbf{u}_q$ with its $(i,j)$-th element as,
\begin{equation}\label{eqn:coefficient}
\mathbf{u}_q({i,j}) = \langle G_q, G_{\sigma_{i,j}}\rangle,
\end{equation}
where $\langle \cdot, \cdot\rangle$ denotes the inner product.
With $\mathbf{u}_q$, spatially variant Gaussian smoothing can then be approximated as,
\begin{equation}\label{eqn:acm}
\widetilde{\mathbf{x}} = \sum_{q=1}^{C} \mathbf{u}_q \circ (\mathbf{x} \ast G_q),
\end{equation}
where $\circ$ denotes entry-wise product, and $\ast$ is convolution operation.
Due to the fact that Gaussian filter is isotropic, $C$ is generally smaller than $(K+1)/2$.
When $N$ is large, Eqn.~(\ref{eqn:acm}) is guaranteed to be a good and efficient approximation.
The efficiency is given more descriptions here.
Take the feature of size $C\times H \times W$ as an example, Eqn.~(\ref{eqn:smoothing}) has to perform pixel-wise matrix multiplication (MM) for each $x_{k,l}$.
In comparison, Eqn.~(\ref{eqn:coefficient}) only need $C\times i$ times MM, by the observation that $G_{\sigma_{i,j}}$ shares the same value for each $j$.
Finally, Eqn.~(\ref{eqn:acm}) only takes about $1/10$ the time compared to Eqn.~(\ref{eqn:coefficient}), making the acceleration up to $\sim$20 times for most cases.
For the degree of approximation, it is measured by the energy preserved.
Please refer to Section.~\ref{sec:training_details} for more details.

There are a few notes about PGC that are worthwhile to mention.
With the spatially variant Gaussian smoothing, PGC can adaptively employ a larger (or smaller) receptive field for people at larger (or smaller) scale, which handles the continuous intra-scene scale variations flexibly.
On the other hand, to effectively enforce the consistency for handling different scales, the conventional spatially invariant convolution is appended in the second stage of PGC, as indicated in Eqn.~(\ref{eqn:persp_conv}).
Furthermore, it is seen from Eqns.~(\ref{eqn:persp_conv}) and (\ref{eqn:acm}) that the complete set of parameters of PGC can be trained end-to-end.

\subsection{Perspective Estimation}
\label{sec:persp_estim}
Since perspective map annotations are seldom available, we introduce a perspective estimation branch to learn the perspective map of an image.
Ideally, with the annotated density map for crowd counting alone, the entire model (including the perspective estimation branch) can be trained end-to-end.
However, such strategy ignores the internal structure of the perspective map and may result in very poor results.
Motivated by \cite{Mostajabi_2018_CVPR}, we suggest a three-phase procedure to train an auto-encoder.

In the first phase, we use perspective maps from WorldExpo'10~\cite{zhang2015cross} to train an auto-encoder $D_{p}(E_{p}(\mathbf{p}; \Theta^E_p); \Theta^D_p)$, which takes a perspective map $\mathbf{p}$ as input and reconstructs it.
$\Theta^E_p$ and $\Theta^D_p$ denote the model parameters of the encoder $E_{p}$ and decoder $D_{p}$, respectively.
%
%Please refer to supplementary material for the details of the encoder-decoder network.
%
The objective is to minimize the $\ell_2$ reconstruction loss,
\begin{equation}\label{eqn:persp_estimating_loss}
{\cal L}_{p2p} = \frac{1}{2N}\sum_{i=1}^{N}\| D_{p}(E_{p}(\mathbf{p}_i; \Theta^E_p);\Theta^D_p) - \mathbf{p}_i \|_{2}^{2} ,
\end{equation}
where $N$ denotes the number of perspective maps.
After training, the latent code $E_{p}(\mathbf{p}; \Theta^E_p)$ can encode the internal structure and contextual relationships of $\mathbf{p}$, while the decoder $D_{p}$ can accurately and robustly recover a high quality perspective map from the latent code.

In the second phase, we use the image and perspective map pairs in WorldExpo'10 to learn another auto-encoder $D_{p}(E_{I}(\mathbf{I}; \Theta^E_I); \Theta^D_p)$, which takes an image $I$ as input to predict its perspective map.
Here, we adopt the decoder parameters $\Theta^D_p$ from the first phase, and only train the encoder parameters $\Theta^E_I$ by minimizing the following loss,
\begin{equation}\label{eqn:persp_perceptual_loss}
{\cal L}_{I2p} = \frac{1}{2N}\sum_{i=1}^{N}\| D_{p}(E_{I}({I}_i; \Theta^E_I);\Theta^D_p) - \mathbf{p}_i \|_{2}^{2} .
\end{equation}

In the third phase, we further train the encoder with another training set.
The encoder is fine-tuned with the loss of the density map if perspective maps are unavailable, and can be optimized by both ${\cal L}_{I2p}$ and density map loss if perspective maps are available.
Benefited from the robustness of the decoder, even if the encoder is not well trained, the decoder can still recover a reasonable perspective map.

Based on the description above, for a new set of training data, our perspective estimation branch can be trained even without corresponding perspective annotations.

\subsection{Network Architecture}
\label{sec:architecture}
% ################# revise this section ####################3
%
Fig. \ref{fig:PGCNet} illustrates the architecture of our PGCNet, which is comprised of three subnetworks, \ie backbone, perspective estimation network (PENet), and density map predictor network (DMPNet).
We adopt CSRNet~\cite{li2018csrnet} as our backbone, where the last convolution (\ie \textit{conv1-1-1}) is removed.
%
% ###### Illustrate this architecture if needed.
The PENet uses the encoder-decoder structure in Sec.~\ref{sec:persp_estim}, which will be described in detail in supplementary material.
As for the DMPNet, in each PGC module, a dilated convolution with factor of 2 is exploited after the spatially variant Gaussian smoothing.
The PGC block is then constructed by concatenating the features before / after the PGC module.
Finally, the outputs from the backbone and PENet are fed to the DMPNet, which stacks five PGC blocks for density map estimation.
%

%%%%%%%%%%%%%%%%%%%%%%%%%%%%%%%%%%%%%%%%%%%%%
% The New Dataset
%
%%%%%%%%%%%%%%%%%%%%%%%%%%%%%%%%%%%%%%%%%%%%%
\section{The Crowd Surveillance Dataset}
\label{sec:new_dataset}
Limited by the annotation difficulty, most public datasets for crowd counting are of relatively small size (as shown in Table~\ref{table:exp_datasets}).
Although larger datasets (\ie WorldExpo'10~\cite{zhang2015cross}) have been proposed, they are nevertheless of low resolution and image quality.
We hence introduce the Crowd Surveillance dataset\protect\footnote{https://ai.baidu.com/broad/subordinate?dataset=crowd\_surv}, which contains 13,945 high-resolution images (386,513 marked people). This means that Crowd Surveillance is nearly 3$\times$ larger than the combination of all the other four datasets in Table~\ref{table:exp_datasets}, leading to the largest dataset with the highest average resolution for crowd counting at present.
Besides, we also provide regions of interest (ROI) annotation for each image to mask out the regions that are too blurry or ambiguous for training / testing.

We build our dataset by both online crawling with search engines and real-life surveillance video acquisition from cooperative partners with necessary permissions.
Fig.~\ref{fig:img_num_bar} provides the statistical histogram of crowd count on different datasets, among which Crowd Surveillance exhibits remarkable high data volume and crowd count variances.
Fig.~\ref{fig:new_dataset} shows the qualitative comparison between our dataset and two most relevant benchmarks, ShanghaiTech Part A~\cite{zhang2016single} and WorldExpo'10~\cite{zhang2015cross}.
From Fig.~\ref{fig:new_dataset} and Table~\ref{table:exp_datasets}, it is seen that although our dataset is less crowded in average density, it provides the highest average resolution (which ensures the image quality), and covers more challenging scenarios with complicated backgrounds and varying crowd count, which significantly increases the difficulty of crowd density estimation. It is also worth noting that although datasets with extremely high crowd densities (such as UCF-QNRF~\cite{idrees2018composition}) have been proposed, they are less suitable for real-world surveillance scenarios, which usually have moderate-to-low crowd density. Our dataset, on the other hand, fits these practical applications well.

\begin{table}
    \begin{center}
    \makebox[\linewidth]{\resizebox{0.85\linewidth}{!}{
      \begin{tabular}{lcccc}
        \toprule
        Dataset & \#Train & \#Test & Avg. Density & Avg. Resolution\\
        \midrule
        ShanghaiTech A & 300 & 182 & 500 & 868$\times$589\\
        ShanghaiTech B & 400 & 316 & 123 & 1,024$\times$768\\
        WorldExpo'10 & 3, 380 & 600 & 56 & 720$\times$576\\
        UCF\_CC\_50 & - & 50 & 1,279 & 902$\times$653\\
        \bf{Crowd Surveillance} & \bf{10,880} & \bf{3,065} & \bf{35} & \bf{1,342$\times$840}\\
        \bottomrule
      \end{tabular}}}
  \end{center}
 % \vspace{-1em}
   \scriptsize
  \caption{Statistics of different crowd counting datasets.}
  \label{table:exp_datasets}
\end{table}

%%%%%%%%%%%%%%%%
%% fig: new_dataset
%%%%%%%%%%%%%%%%
\vspace{-1em}
\begin{figure}
\centering
\includegraphics[width=0.4\textwidth]{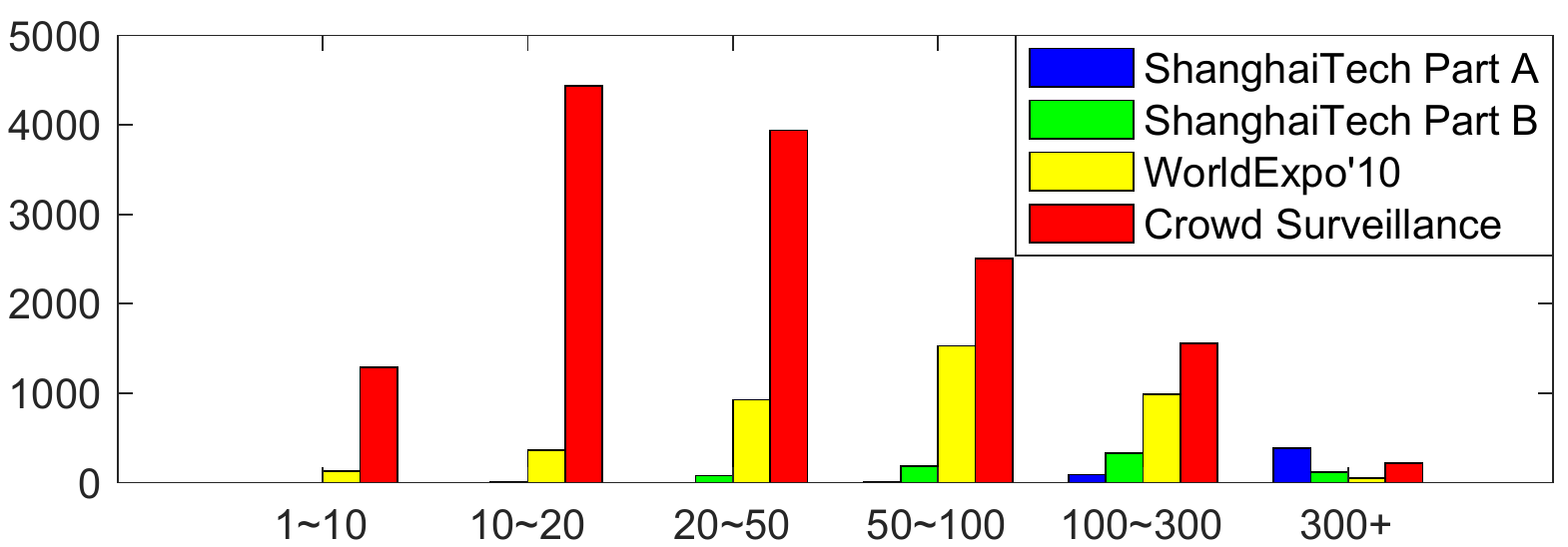}
%\vspace{-1em}
 \scriptsize
\caption{Statistical histogram of crowd counts on different datasets. It is seen that our dataset exhibits high data volume and crowd count variance.}
\label{fig:img_num_bar}
\end{figure}

\begin{figure}
\centering
\includegraphics[width=0.4\textwidth]{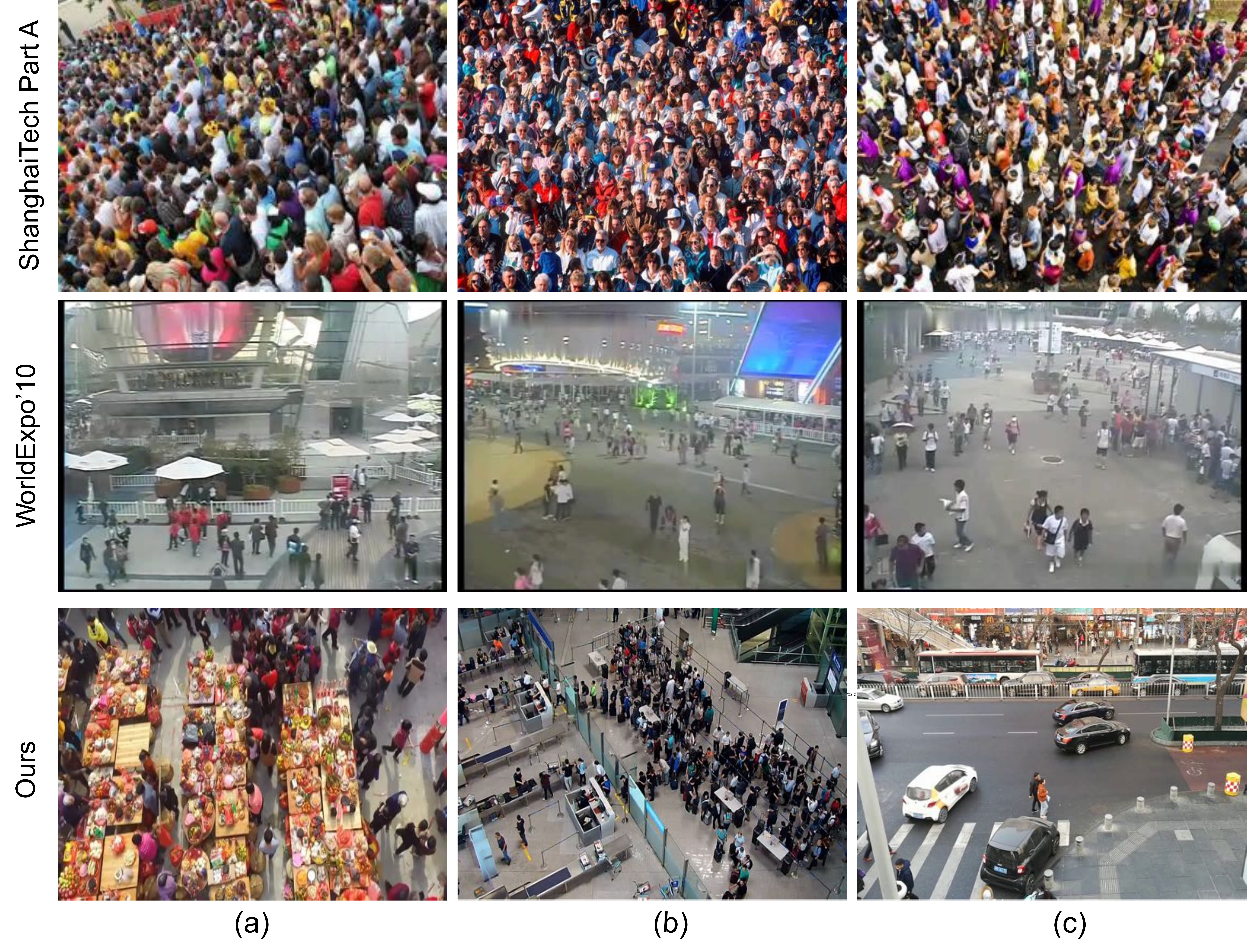}
%\vspace{-1em}
 \scriptsize
\caption{Qualitative comparison of crowd counting datasets. Our dataset is seen to have more challenging scenarios and higher resolutions than existing datasets.}
\label{fig:new_dataset}
\end{figure}

%%%%%%%%%%%%%%%%%%%%%%%%%%%%%%%%%%%%%%%%%%%%%
% Experimental Results
%
%%%%%%%%%%%%%%%%%%%%%%%%%%%%%%%%%%%%%%%%%%%%%
\section{Experimental Results}
\label{sec:experiments_results}
In this section, we first present our implementation details, and then compare the proposed PGCNet with the state-of-the-arts on four public datasets, namely ShanghaiTech~\cite{zhang2016single}, WorldExpo'10~\cite{zhang2015cross}, UCF\_CC\_50~\cite{idrees2013multi} and our proposed Crowd Surveillance. Extensive ablation study is then conducted to reveal the contribution of each component in PGCNet.
We adopt mean absolute error (MAE) and mean square error (MSE) as metrics for evaluating crowd counting and perspective estimation.
%
%
%\subsection{Evaluation Metrics}
%\label{sec:eval_metric}
%Following previous works~\cite{zhang2015cross, zhang2016single, sam2017switching}, we adopt mean absolute error (MAE) and mean square error (MSE) as our evaluation metrics.

%, which are defined as $MAE=\frac{1}{N}\sum_{i=1}^{N}{|C_{i}-C_{i}^{GT}|}$ and $MSE=\sqrt{\frac{1}{N}\sum_{i}^{N}\|C_{i}-C_{i}^{GT}\|^2}$
%\begin{equation}\label{MAE_MSE}
%\begin{split}
%& MAE=\frac{1}{N}\sum_{i=1}^{N}{|C_{i}-C_{i}^{GT}|} \\
%& MSE=\sqrt{\frac{1}{N}\sum_{i}^{N}\|C_{i}-C_{i}^{GT}\|^2} \\
%\end{split}
%\end{equation}
%where $N$ is the testing image number, $C_i$ is the estimated crowd count, and $G_i^{GT}$ is the ground-truth count.
%
%Similarly, the MAE and MSE for perspective estimation can be defined by replacing $C_i$ and $C_{i}^{GT}$ with $P_i$ and $P_{i}^{GT}$ in Eqn.~(\ref{MAE_MSE}), where $P_i$ is the perspective estimation and $P_i^{GT}$ is
%the ground-truth perspective value.
%

\subsection{Implementation Details}
\label{sec:training_details}
Denote by $\Theta$ the model parameters of the full PGCNet $\Phi(I; \Theta)$.
Given the training data $\{(I_i, Y_i)\}$, the network can be trained by minimizing the following objective function,
\begin{equation}\label{eqn:pmae_pmse}
L\left(\Theta\right) = \frac{1}{2N}\sum_{i=1}^{N}\|\Phi(I; \Theta) - Y_{i}\|_{2}^{2},
\vspace{-0.5em}
\end{equation}
where $Y_i$ is the ground-truth density map of $I_i$.
If perspective map annotations are also available, we can further incorporate ${\cal L}_{I2p}$ in Eqn.~(\ref{eqn:persp_perceptual_loss}) for better training of the PENet.

To train PGCNet, we adopt stochastic gradient descent (SGD) with fixed learning rate $10^{-7}$ and weight decay $5\times10^{-4}$.
The momentum is set to 0.95 and the batch size is set to 1.
PGCNet takes a whole image as input during both training and testing.
Suppose there is a dot annotation of people head at location $(x_i, y_i)$ represented by a delta function $\delta(x-x_i)\delta(y-y_i)$, the ground-truth density map $Y$ is obtained by convolving each annotation point with a normalized Gaussian kernel $G_{\sigma}$: $Y = \sum_{i}^{S}\delta(x-x_i)\delta(y-y_i)*G_{\sigma}$, where $S$ is the total number of dot annotations in the image, and $\sigma$ is set to fixed value 0.5.

To establish the baseline, we reimplement CSRNet~\cite{li2018csrnet} on Pytorch~\cite{paszke2017automatic} (denoted as CSRNet*) with the four datasets adopted, which is expected to deliver comparable performances against~\cite{li2018csrnet}.
The weights of the PGC modules are initialized with Gaussian distribution of zero mean and 0.01 standard deviation, while the other layers are initialized with the corresponding pre-trained weights from CSRNet*.
Random flipping is adopted for data augmentation.
For ShanghaiTech and WorldExpo'10, since their ground-truth perspective maps are available, they are used as the guidance of PGC directly, and our model is trained for 300 epochs without PENet.
%
% #### rewrite these words
For the datasets without perspective annotations, the PENet is pre-trained for the first two phases of Sec.~\ref{sec:persp_estim}, both 500 epochs; the backbone, PENet and DMPNet are then trained together for another 300 epochs, as the third phase of Sec.~\ref{sec:persp_estim}.

The range of $\sigma$ we sample is $[1/4, K/4]$ with $P\{\|X-\mu \| < 2\sigma\}\approx 0.95$.
We pre-define a group of Gaussian candidates $G_{\sigma}$ of size $K\times K$ with $\sigma$ ranging in $\left[1/4, K/4\right]$ on the step $s$.
$K$ is set to 7 by default, unless explicitly stated; and  $s$ is set to $0.05$ since no significant gain is observed with denser steps.
The total number of Gaussian candidates is $N=30$ with $\sigma$ sampling in $\left[1/4, 7/4\right]$ on step $s=0.05$.
PCA is then applied on $G_{\sigma}$ to get the $(K+1)/2$ eigenvectors $G_q$ in Eqn.~(\ref{eqn:coefficient}).
In this case, the original matrix is $30\times7\times7$, which will be reshaped to $30\times49$ and applied PCA to get the approximate matrix $4\times49$.
The energy preserved is $99.9\%$, which is the ratio between the sum of singular values of approximated matrix against those of the original matrix.
$\alpha$ and $\beta$ are initialized by normalizing the $\mathbf{p}$ in Eqn.~(\ref{eqn:ada_sig}) to $(0,1)$, while
$p_0$ and $a$ are empirically set to 0 and 1.
All the parameters above are differentiable and trainable.
As a tradeoff between efficiency and performance, our results are achieved by stacking five PGC blocks for all datasets.
It takes about 2 days to train the network on WolrdExpo'10 with an NVIDIA Tesla P40 GPU.

\subsection{Evaluations and Comparisons}
\label{sec:eval_and_comparison}
Four datasets are adopted in our experiments, including ShanghaiTech~\cite{zhang2016single}, WorldExpo'10~\cite{zhang2015cross}, UCF\_CC\_50~\cite{idrees2013multi} and the proposed Crowd Surveillance.
For the later two that do not have perspective map annotations, we denote \emph{Ours A} as directly adopting the estimated perspective maps (based on the PENet trained on perspective annotations from ShanghaiTech A) as \emph{ground-truth} and feed them to the training of PGC;
while \emph{Ours B} as the end-to-end training without perspective map annotations described in Sec.~\ref{sec:persp_estim}, where the backbone, PENet, and DMPNet are jointly trained without any density map annotation.
For the estimated perspective maps of both \emph{Ours A} and \emph{Ours B}, we use the mean of each line to replace the values of the whole line, which forms the \emph{final} estimated perspective map.
For \emph{Ours A}, by following Sec.~\ref{sec:persp_estim} with ShanghaiTech A, we get 0.020 MAE and 0.031 MSE for the first phase of perspective map estimation, and 0.101 MAE and 0.142 MSE for the second phase.
Visualization of perspective estimation and more details of training of the PENet will be illustrated in our supplementary materials.

\noindent{\textbf{ShanghaiTech}} contains 1,198 images with a total of 330,165 annotated people.
The dataset is split into Part A and Part B, with 482 and 716 images respectively.

We adopt the ground-truth perspective maps provided by~\cite{shi18pacnn} as the guidance of DMPNet.
The results are listed in Table~\ref{table:shanghai}. It is seen that our method and SANet~\cite{Cao2018Scale} dominate the top ranks, where ours achieves the best result on Part A \wrt both MAE and MSE with significant margins, and is only slightly surpassed by SANet on Part B. Our method also exhibits significant performance gain over the baseline CSRNet*.
Besides, some test cases can be found in Fig.~\ref{fig:sample}, clearly indicating that the superiority of PGCNet over CSRNet in estimating a better density map.

%%%%%%%%%%%%%%%%%%%%%%%%%%%%%%%%%%%%%%%%%%%%%%%%%%%%%%%%%%%%%%%%%%%%%
%
%  Sample Images of PGCNet
%
%%%%%%%%%%%%%%%%%%%%%%%%%%%%%%%%%%%%%%%%%%%%%%%%%%%%%%%%%%%%%%%%%%%%%
\begin{figure}[!t]
\centering
\includegraphics[width=0.48\textwidth]{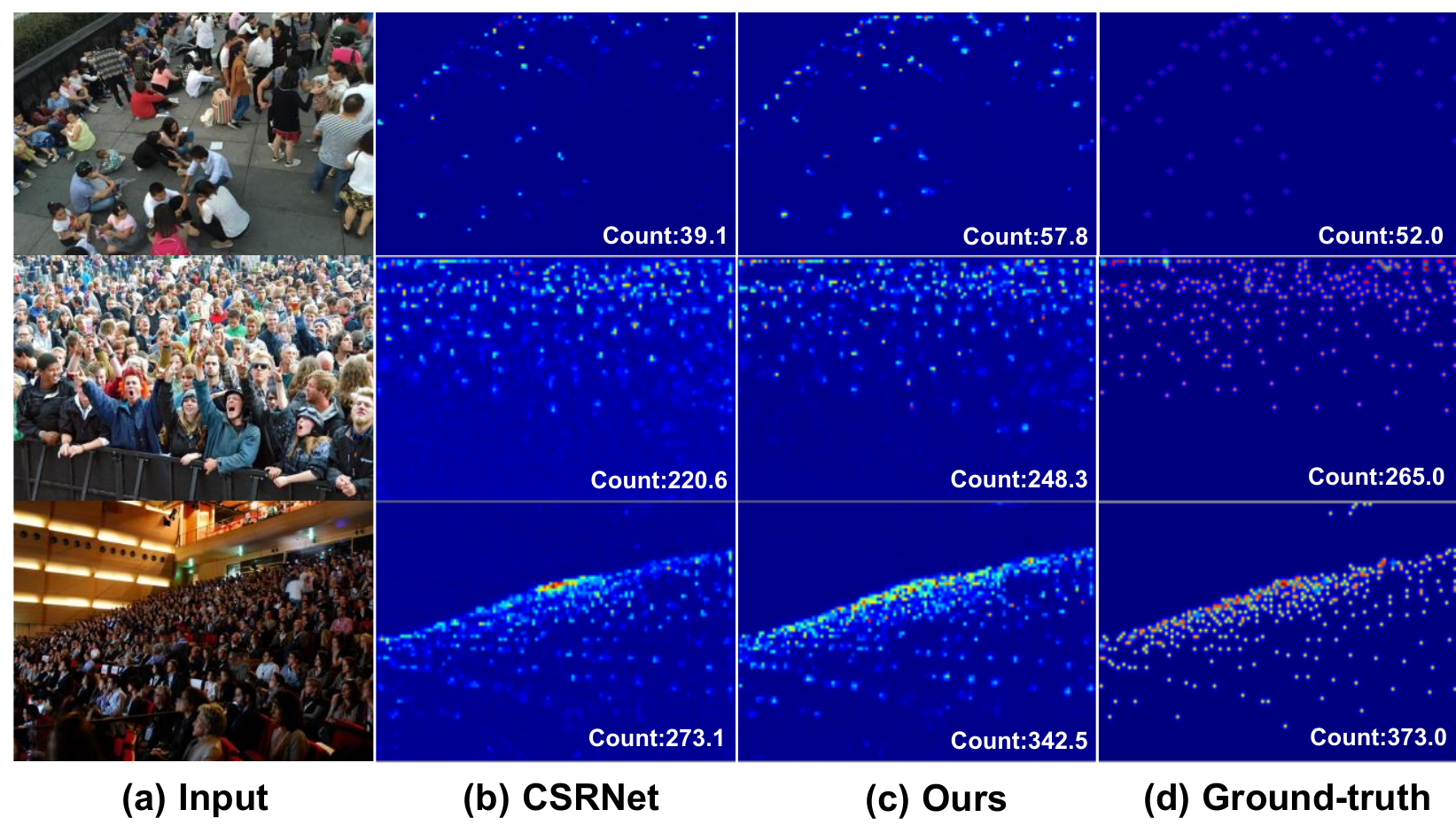}
  \vspace{-1.5em}
 \scriptsize
\caption{Density maps estimated by CSRNet~\cite{li2018csrnet} and Ours.}
\label{fig:sample}

\end{figure}

\begin{table}[!h]
   \begin{center}
    \resizebox{0.85\hsize}{!}{
      \begin{tabular}{| c | c  c | c  c|}
      %\toprule
      \hline
      \multirow{2}{*}{Method}& \multicolumn{2}{c|}{Part A} & \multicolumn{2}{c|}{Part B} \\
      \cline{2-5}
      %\midrule
      & MAE & MSE & MAE & MSE \\
      \hline

     % \hline
      Zhang~\etal~\cite{zhang2015cross} & 181.8 & 277.7 & 32.0 & 49.8 \\
     % \hline
      MCNN~\cite{zhang2016single} & 110.2 & 173.2 & 26.4 & 41.3 \\
    %  \hline
      Cascaded-MTL~\cite{boominathan2016crowdnet} & 101.3 & 152.4 & 20.0 & 31.1 \\
    %  \hline
      Switching-CNN~\cite{sam2017switching} & 90.4 & 135.0 & 21.6 & 33.4 \\
    %  \hline
      CP-CNN~\cite{sindagi2017generating} & 73.6 & 106.4 & 20.1 & 30.1 \\
     % \hline
      PACNN~\cite{shi18pacnn} & 84.5 & 132.5 & 14.2 & 24.1 \\
   %   \hline
      DecideNet~\cite{liu2018decidenet} & - & - & 20.8  &  29.4 \\
     % \hline
      SANet~\cite{Cao2018Scale} & 67.0 & 104.5 & \textbf{8.4} & \textbf{13.6} \\
    %  \hline
      CSRNet~\cite{li2018csrnet} & 68.2 & 115.0 & 10.6 & 16.0 \\
    %  \hline
      CSRNet* & 67.5 & 103.1 & 10.7 & 16.4\\
    %  \hline
      Ours & \textbf{57.0} & \textbf{86.0} & 8.8 & 13.7 \\
    %\bottomrule
    \hline
  	\end{tabular}
  	}
  	\end{center}
 \vspace{-1em}\
 \scriptsize
\caption{Comparisons on ShanghaiTech dataset~\cite{zhang2016single}.}
\label{table:shanghai}
\end{table}

\noindent\textbf{WorldExpo'10} contains 3,980 images from the 2010 Shanghai WorldExpo.
The training set contains 3,380 images, while the test set is divided into five different scenes with 120 images each.
ROIs are provided to indicate the target regions for training / testing.
Following~\cite{li2018csrnet}, each image and its ground-truth density map are masked with the ROI in preprocessing.
We use the official ground-truth perspective map to guide the processing of PGC.
The results are shown in Table~\ref{table:WorldExpo}, where our method achieves the best 8.1 average MAE against other methods.

\begin{table}[!h]
   \begin{center}
    \resizebox{1\hsize}{!}{
      \begin{tabular}{| c | c  c  c  c  c  c |}
      %\toprule
      \hline
      Method & Sce.1 & Sce.2 & Sce.3 & Sce.4 & Sce.5 & Avg. \\
      %\midrule
      \hline
      Zhang~\etal~\cite{zhang2015cross} & 9.6 & 14.1 & 14.3 & 22.2 & 3.7 & 12.9 \\
      %\hline
      MCNN~\cite{zhang2016single} & 3.4 & 20.6 & 12.9 & 13.0 & 8.1 & 11.6 \\
     %\hline
      Switching-CNN~\cite{sam2017switching} & 4.4 & 15.7 & 10.0 & 11.0 & 5.9 & 9.4 \\
      %\hline
      CP-CNN~\cite{sindagi2017generating} & 2.9 & 14.7 & 10.5 & 10.4 & 5.8 & 8.9 \\
      %\hline
      PACNN~\cite{shi18pacnn} & 2.6 & 15.4 & 10.6 & 10.0 & 4.8 & 8.7 \\
      %\hline
      DecideNet~\cite{liu2018decidenet} & 2.0 & 13.1 & 8.9 & 17.4 & 4.8 & 9.2 \\
      % \hline
      SANet~\cite{Cao2018Scale} & 2.6 & 13.2 & 9.0 & 13.3 & 3.0 & 8.2 \\
     % \hline
      CSRNet~\cite{li2018csrnet} & 2.9 & 11.5 & 8.6 & 16.6 & 3.4 & 8.6 \\
      %\hline
      CSRNet* & 2.4 & 15.1 & 7.9 & 15.6 & 2.7 & 8.7 \\
      %\hline
      Ours & 2.5 & 12.7 & 8.4 & 13.7 & 3.2 & \textbf{8.1} \\
      %\bottomrule
      \hline
  	\end{tabular}
  	}
  	\end{center}
	  \vspace{-1em}
      \scriptsize
	  \caption{Comparisons on WorldExpo'10~\cite{zhang2015cross} dataset.}
\label{table:WorldExpo}
\end{table}

\noindent\textbf{UCF\_CC\_50} contains 50 images of diverse scenes.
The head count per image varies drastically (from 94 to 4,543).
Following~\cite{idrees2013multi}, we split the dataset into five subsets and perform a 5-fold cross-validation.
Since the perspective map annotation is unavailable, we conduct the two comparison experiments \emph{Ours A} and \emph{Ours B} described at the beginning of this section.
The results are shown in Table~\ref{table:UCFCC}, where our method achieves the optimal performance with large margins. Compared with the baseline CSRNet*, \emph{Ours A} achieves significant gain on both MAE and MSE.

We further note that \emph{Ours B} achieves 244.6 MAE, with another 14.8 gain over \emph{Ours A}, which shows the feasibility of our end-to-end training strategy described in Sec.~\ref{sec:persp_estim}.
\begin{table}[!h]
   \begin{center}
    \resizebox{0.6\hsize}{!}{
      \begin{tabular}{| c | c  c |}
      %\toprule
      \hline
      Method & MAE & MSE \\
      %\midrule
      \hline
      Idrees~\etal~\cite{idrees2013multi} & 419.5 & 541.6 \\
      %\hline
      Zhang~\etal~\cite{zhang2015cross} & 467.0 & 498.5 \\
      %\hline
      MCNN~\cite{zhang2016single} & 377.6 & 509.1  \\
     % \hline
      Cascaded-MTL~\cite{boominathan2016crowdnet} & 322.8 & 341.4  \\
     % \hline
      Switching-CNN~\cite{sam2017switching} & 318.1 & 439.2  \\
     % \hline
      CP-CNN~\cite{sindagi2017generating} & 295.8 & 320.9  \\
      % \hline
      PACNN~\cite{shi18pacnn} & 304.9 & 411.7 \\
      % \hline
      SANet~\cite{Cao2018Scale} & 258.4 & 334.9 \\
      %\hline
      CSRNet~\cite{li2018csrnet} & 266.1 & 397.5  \\
     % \hline
      CSRNet* & 264.0 & 398.1 \\
      %\hline
      Ours A & 259.4 & \textbf{317.6}  \\
      %\hline
      Ours B & \textbf{244.6} & 361.2  \\
      %\bottomrule
      \hline
  	\end{tabular}
  	}
  	\end{center}
	  \vspace{-1em}
        \scriptsize
        \caption{Comparisons on UCF\_CC\_50~\cite{idrees2013multi} dataset.}
\label{table:UCFCC}
\end{table}

\noindent\textbf{Crowd Surveillance}\label{sec:crowd_surveillance} is our newly proposed dataset with 10,880 and 3,065 images for training / testing, as illustrated in Sec.~\ref{sec:new_dataset}.
Similar to WorldExpo'10, we also provide ROI annotations that are used in the preprocessing.
Table~\ref{table:Crowd_Surveillance} demonstrates the comparisons of our method against MCNN~\cite{zhang2016single}, Switching-CNN~\cite{sam2017switching} and CSRNet~\cite{li2018csrnet}, and our method achieves the best results.
Moreover, when we directly adopt the perspective map estimated by the PENet pre-trained on ShanghaiTech A (\emph{Ours A}), we achieve 2.1 MAE gain over the baseline; and when we train the model end-to-end (\emph{Ours B}), another 0.5 MAE gain over \emph{Ours A} is achieved.
We also provide visualization results of a test example in Fig.~\ref{fig:mul_stage}.
It is seen that even perspective annotation is completely absent in Crowd Surveillance, the PENet is still able to provide reasonable perspective estimations, which supports its generalization ability.
Such observations further validates our end-to-end training strategy described in Sec.~\ref{sec:persp_estim}.
%%%%%%%%%%%%%%%%%%%%%%%%%%%%%%%%%%%%%%%%%%%%%%%%%%%%%%%%%%%%%%%%%%%%%
%
%  Comparisons of multi stage
%
%%%%%%%%%%%%%%%%%%%%%%%%%%%%%%%%%%%%%%%%%%%%%%%%%%%%%%%%%%%%%%%%%%%%%
\begin{figure}[!t]
\centering
\includegraphics[width=0.5\textwidth]{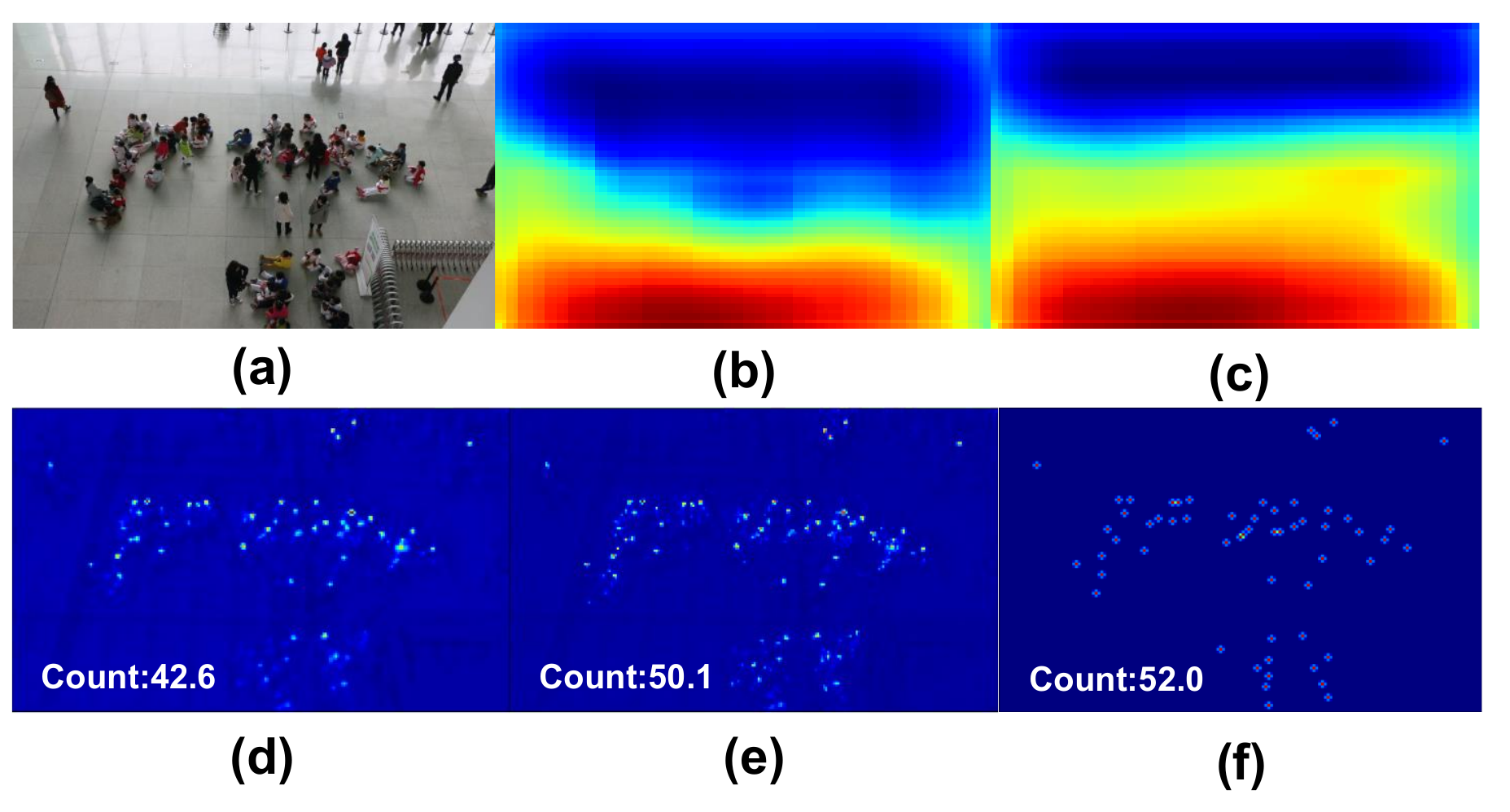}
 \scriptsize
\caption{Visualization of a test example from Crowd Surveillance. (a) is the input; (b)(c) / (d)(f) are estimated perspective / density maps of Ours A and Ours B; and (f) is the ground truth.}
\label{fig:mul_stage}
\vspace{-1em}
\end{figure}

\begin{table}[!h]
   \begin{center}
    \resizebox{0.6\hsize}{!}{
      \begin{tabular}{| c | c  c |}
     %\toprule
     \hline
      Method & MAE & MSE \\
      %\midrule
      \hline
      MCNN~\cite{zhang2016single} & 23.8 & 49.9  \\
     % \hline
      Switching-CNN~\cite{sam2017switching} & 16.9 & 33.2  \\
      %\hline
      CSRNet* & 9.8 & 21.6 \\
      %\hline
      Ours A & 7.7 & 16.4  \\
     % \hline
      Ours B & \textbf{7.2} & \textbf{15.6}  \\
      %\bottomrule
      \hline
  	\end{tabular}
  	}
  	\end{center}
	  \vspace{-1em}
     \scriptsize
	\caption{Comparisons on Crowd Surveillance.}
\label{table:Crowd_Surveillance}
\end{table}

\subsection{Ablation Study}
\label{sec:ablation}

The evaluations against the state-of-the-arts above demonstrate the superiority of our PGC block and the end-to-end training strategy.
We first conduct the experiment on choosing the appropriate value of $K$.
Then we show the influence of the number of PGC blocks stacked in our network.
%
% (Remove it to supp)
% Besides, we confirm the effectiveness of the PGC by replacing it with two types of feature concatenations.
%
Moreover, we demonstrate the feasibility of PGC block being an insertable component for existing network to improve performance.
Finally, we also verify the importance of the pre-training of PENet in our method.

\subsubsection{Influence of $K$}
\label{sec:value_K}
Table~\ref{table:comparison_K} lists the results of our method with a single PGC block but different $K$, \ie the Gaussian filter size of Sec.~\ref{sec:persp_conv}.
When $K$ is small (\eg $\le 5$), the PGC module is computationally light ($\sim$6ms for a $576\times720$ image, similarly hereinafter) but has poor performance;
when $K$ grows too large (\eg $\geq 9$), the performance starts to degrade with heavier computational burden ($\sim$16ms);
when $K=7$, it performs optimally with affordable efficiency ($\sim$10ms).
We therefore adopt $K=7$ in our following experiments.
\begin{table}[!h]
   \begin{center}
    \resizebox{0.85\hsize}{!}{
      \begin{tabular}{ |c| c c | c| c c| }
      \hline
       $K$& MAE & MSE & $K$ & MAE & MSE \\
      \hline
      $K=1$ & 67.3 & 100.0 & $K=7$ & \textbf{65.8} & 98.0 \\
      %\hline
      $K=3$ & 66.2 & \textbf{97.7} & $K=9$ & 66.3 & 98.8 \\
      %\hline
      $K=5$ & 66.4 & 98.3 & $K=11$ & 66.4 & 97.4 \\
      \hline
  	\end{tabular}
  	}
  	\end{center}
%	  \vspace{-1em}
     \scriptsize
	 \caption{Influence of $K$ on a single PGC block on ShanghaiTech Part A.}
\label{table:comparison_K}
\end{table}

%\vspace{-1em}

%

\subsubsection{Influence of the Number of PGC Blocks}
\label{sec:number_of_persp_convs}
%
%%%%%%%%%%%%%%%%%%%%%%%%%%%%%%%%%%%%%%%%%%%%%%%%%%%%%%%%%%%%%%%%%%%%%
%
%  Visualization of multi pgc
%
%%%%%%%%%%%%%%%%%%%%%%%%%%%%%%%%%%%%%%%%%%%%%%%%%%%%%%%%%%%%%%%%%%%%%
\begin{figure}[!t]
\centering
\includegraphics[width=0.47\textwidth]{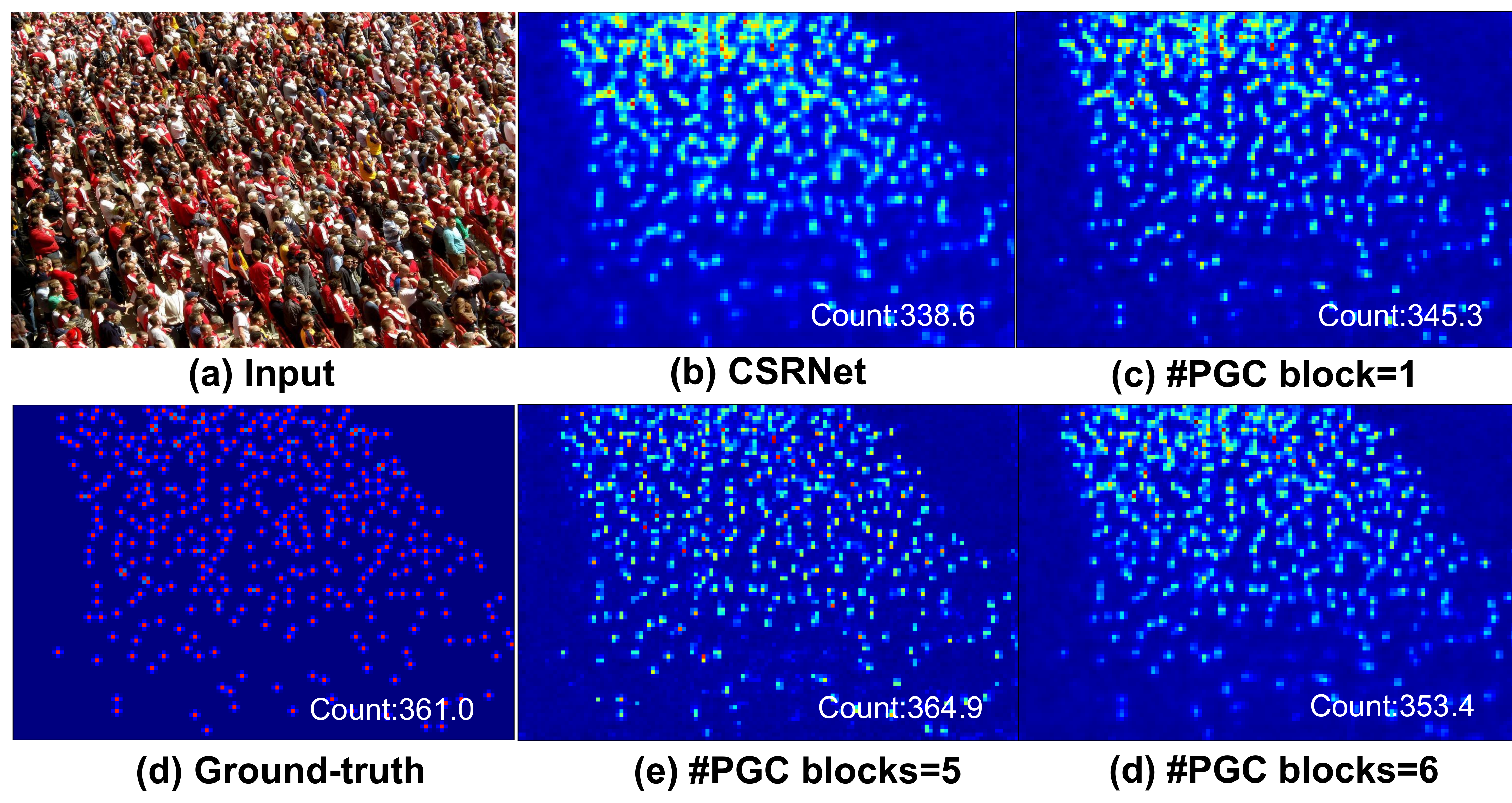}
 %\vspace{-1em}
\scriptsize
\caption{Density maps predicted by stacking of different number of PGC blocks.}
\label{fig:mul_pgc}
\end{figure}

%%%%%%%%%%%%%%%%%%%%%%%%%%%%%%%%%%%%%%%%%%%%%%%%%%%%%%%%%%%%%%%%%%%%%
%
%  Comparisons of multi pgc
%
%%%%%%%%%%%%%%%%%%%%%%%%%%%%%%%%%%%%%%%%%%%%%%%%%%%%%%%%%%%%%%%%%%%%%

\begin{table}[!h]
%\vspace{-1em}
   \begin{center}
    \resizebox{0.8\hsize}{!}{
      \begin{tabular}{ |c | c c |c c|}
      %\toprule
      \hline
      \multirow{2}{*}{\# of blocks} & \multicolumn{2}{c|}{Part A} & \multicolumn{2}{c|}{Part B} \\
      %\midrule
      \cline{2-5}
      & MAE & MSE & MAE & MSE \\
      \hline
      1 & 65.8 & 98.0 & 9.8 & 15.8 \\
      %\hline
      2 & 64.5 & 96.6 & 9.6 & 15.4 \\
      %\hline
      3 & 60.9 & 95.2 & 9.2 & 14.9 \\
      %\hline
      4 & 58.5 & 89.5 & 9.1 & 14.4 \\
      %\hline
      5 & \textbf{57.0} & \textbf{86.0} & \textbf{8.8} & \textbf{13.7} \\
     % \hline
      6 & 58.3 & 90.2 & 9.0 &  14.2 \\
      %\bottomrule
      \hline
  	\end{tabular}
  	}
  	\end{center}
	  \vspace{-1em}
      \scriptsize
	 \caption{Influence of the number of stacked PGC blocks in our method on ShanghaiTech.}
\label{table:differnet_modules}
\end{table}
Table~\ref{table:differnet_modules} shows the performance of our network when stacking different numbers of PGC blocks, as Fig.~\ref{fig:PGCNet} shows.
The performance increases with the number of PGC blocks stacked until reaching peak values (57.0 and 8.8 MAE on Part A and Part B, respectively) with 5 blocks, and degrades afterwards.
Fig.~\ref{fig:mul_pgc} demonstrates the predicted density maps when stacking serveral PGC blocks.
It is seen that PGC gradually refines the density map till 5 PGC blocks.
Too many PGC blocks(\eg, 6) may lead to over-smoothing of features, resulting in worse estimation of density maps.
We hence determine our final network architecture with 5 PGC blocks stacked, resulting in 100ms total time cost per 576$\times$720 image in testing.

\subsubsection{Extensibility to Another Backbone}
\label{ResNet_with_PGC_blocks}
In order to verify the extensibility of our PGC block to another backbone instead of VGG-16~\cite{simonyan2014very} in CSRNet*, we adopt a truncated ResNet-101~\cite{he2016deep} with the first 10 convolutional layers.
It then appends PGC blocks (three in our experiment) and three extra convolutional layers are added afterwards to reduce the channel dimension to 1, making it compatible with the density map.
Table~\ref{table:res_pgc} shows the comparison results on ShanghaiTech Part A / B, where \emph{ResNet-101(backbone)} is the backbone itself, and \emph{ResNet-101(PGC)} appends three PGC blocks.
All models are trained for 500 epochs with the first 10 convolutional layers initialized by pre-trained weights on ImageNet~\cite{deng2009imagenet}.
For \emph{ResNet-101(backbone)}, we get 109.6/26.2 MAE and 187.6/40.5 MSE.
It is seen that when we stack three PGC blocks (\emph{ResNet-101(PGC)}), we obtain the most performance gain of 19.9/7.6 MAE against \emph{ResNet-101(backbone)}, reaching 89.7/18.6 MAE.
Such observations validate the effectiveness and extensibility of the PGC block on another backbone.
\begin{table}[!h]
   \begin{center}
    \resizebox{1.0\hsize}{!}{
      \begin{tabular}{ |c | c  c | c  c|}
      %\toprule
      \hline
      \multirow{2}{*}{Method} & \multicolumn{2}{c|}{Part A} & \multicolumn{2}{c|}{Part B} \\
      %\midrule、
      \cline{2-5}
       & MAE & MSE & MAE & MSE \\
     \hline
      ResNet-101(backbone) & 109.6 & 187.6 & 26.2 & 40.5 \\
    %  \hline
      ResNet-101(PGC) & \textbf{89.7} & \textbf{148.4} & \textbf{18.6} & \textbf{30.9} \\
      \hline
  	\end{tabular}
  	}
  	\end{center}
 \vspace{-1em}
  \scriptsize
\caption{Comparisons on the ResNet-101 backbone.}
\label{table:res_pgc}
\end{table}

\vspace{-1em}
\subsubsection{Importance of the pre-training of PENet}
\label{sec:Importance_of_pre-training}
Although the end-to-end training strategy has been proposed, we still note that the PENet requires fair pre-training for accurate perspective estimation.
To validate the necessity, we conduct two experiments without pre-training of PENet on UCF\_CC\_50~\cite{idrees2013multi} and Crowd Surveillance, and compare the results in Table.~\ref{table:UCFCC} and Table.~\ref{table:Crowd_Surveillance}.
As shown in Table~\ref{table:PENet_pretraining}, the results with PENet pre-training significantly outperform those without pre-training.
Such performance margin is attributed to the confusing guidance of PENet on the spatially variant Gaussian smoothing of Sec.~\ref{sec:persp_conv}, since without pre-training, we observe that the output of PENet is still messy even after considerable epochs of training.

\begin{table}[!h]
	\begin{center}
		\resizebox{1.0\hsize}{!}{
			\begin{tabular}{ |c|  c c| c c|}
				%\toprule
				\hline
				\multirow{2}{*}{Method} & \multicolumn{2}{c|}{UCF\_CC\_50} & \multicolumn{2}{c|}{Crowd Surveillance} \\
				%\midrule
				\cline{2-5}
				& MAE & MSE & MAE & MSE \\
				\hline
				PENet(w/ pre-training) & \textbf{244.6} & \textbf{361.2} & \textbf{7.2} & \textbf{15.6} \\
				%\hline
				PENet(w/o pre-training) & 278.6 & 403.5 & 10.3 & 24.7 \\
				%\bottomrule
				\hline
			\end{tabular}
		}
	\end{center}
	\vspace{-1em}
     \scriptsize
	\caption{The performances with / without pre-training of PENet on UCF\_CC\_50~\cite{idrees2013multi} and Crowd Surveillance.}
	\label{table:PENet_pretraining}
\end{table}

%%%%%%%%%%%%%%%%%%%%%%%%%%%%%%%%%%%%%%%%%%%%%
% Conclusion
%
%%%%%%%%%%%%%%%%%%%%%%%%%%%%%%%%%%%%%%%%%%%%%
\vspace{-1em}
\section{Conclusion}
\label{sec:conclusion}
In this paper, we present a perspective-guided convolution network (PGCNet) for crowd counting.
The key idea of PGCNet is the perspective-guided convolution, which functions as an insertable module that successfully handles the continuous intra-scene scale variation issue.
We also propose a perspective estimation branch as well as its learning strategy, which is incorporated into our method to form an end-to-end trainable network, even without perspective map annotations.
A new large scale dataset Crowd Surveillance is introduced as well to promote the researches in crowd counting.
Experiments on four benchmark datasets show the superiority of our PGCNet against the state-of-the-arts.

\subsubsection*{Acknowledgement}
This work was partially supported by the National Natural Science Foundation of China under grant No. 61671182.

\clearpage

{\small
\bibliographystyle{ieee_fullname}
\bibliography{egbib}
}

\clearpage

\begin{appendix}

\section*{Supplementary Material}

The following items are included in the supplementary materials:

\begin{itemize}
  \item The architecture of PENet and the details of three training phases of PENet.
  \item The visualization of estimated perspective maps in each phase of training PENet.
  \item Reliability of the prediction of PENet.
  \item More density maps predicted by the proposed PGCNet.
\end{itemize}

%%%%%%%%%%%%%%%%%%%%%%%%%%%%%%%
%       Sec: The Architecture of PENet and the Training Details
%%%%%%%%%%%%%%%%%%%%%%%%%%%%%%%
\section{The Architecture of PENet and the Training Details}
\label{sec:PENet}
We adopt \emph{Convolution-LeakyReLU} as the basic pattern of the encoder $E_p$, with each block scaling down the encoder feature by ratio 2.
While the decoder enlarges the resolution of encoded feature by the combination of \emph{UpConv-ReLU}.
Details of the architecture of PENet are demonstrated in Table~\ref{table:PENet}.

In the first phase, PENet is trained to reconstruct the input when given certain perspective map.
As PGCNet adopts CSRNet as the baseline, the resolution of perspective needed by PGC block is only $1/8$ size of the original input.
Therefore, we do not need to predict the full resolution of perspective map.
We downsample the original image to make the resized image be only $1/8$ resolution of the original image and then train PENet as an identity mapping of perspective maps.
We get 0.020 MAE and 0.031 MSE in this phase.

In the second stage, we fix the parameters of $D_p$ trained in the first phase, and only train $E_p$, aiming
at constructing the perspective map from its corresponding RGB image.
We get 0.101 MAE and 0.142 MSE in the second training phase.
And finally in the third stage, \emph{Ours A} denotes directly adopting the estimated perspective map of PENet as the ground-truth,
while \emph{Ours B} represents the PENet is embedded as a perspective estimation branch and the whole network can be trained end-to-end.
Quantitative results of \emph{Ours A} and \emph{Ours B} have been demonstrated in Sec.~5.3.
%

%%%%%%%%%%%%%%%%%%%%%%%%%%%%
%       Sec: Visualization of Estimated Perspective Maps
%%%%%%%%%%%%%%%%%%%%%%%%%%%%
\section{The Visualization of Estimated Perspective Maps in Each Phase of Training PENet}
Beyond the quantitative results given in Sec.~\ref{sec:PENet}, we also demonstrate visualizations of perspective maps in these three phases.
Fig.~\ref{fig:d2d}(a)(b) show the input / ground-truth and estimated perspective map, respectively.
It can be seen that our PENet performs well in reconstructing the input in the first phase.
Fig.~\ref{fig:rgb2d} demonstrates two examples of estimated maps predicted by PENet.
PENet generally produces roughly accurate perspective maps, showing its robustness in dealing with different scenes.
Taking into account the quantitative results, it is obvious that PENet is capable of predicting a meaningful perspective map quantitatively and qualitatively in the first two stages.
%

% The third phase
For the third phase, Fig.~\ref{fig:phase3}  shows the estimated perspective maps of \emph{Ours A} and \emph{Ours B}, respectively shown in the second and third columns.
Comparing these two images in Fig.~\ref{fig:phase3}(a) vertically, it can be seen that the visual angle of the image in the second row is relatively larger than that of the image
in the first row.
This observation accords with the directly estimated maps in the Fig.~\ref{fig:phase3}(b), which can been seen that the second image contains more larger values comparing with
the first image does.
When we train the whole network end-to-end, the perspective estimation branch can still predict generally satisfying perspective maps (\ie, Fig.~\ref{fig:phase3}(c)).
It is seen that larger perspective values move from the right to the left, which is visually explanatory.

Therefore, our PENet works well either in directly predicting perspective maps or in functioning as the perspective map estimator of the end-to-end architecture.

% ##################################################
%       Sec: Reliability of the prediction of PENet
% ##################################################
\section{Reliability of the Prediction of PENet}
\label{sec:Reliability_of_PENet}
PENet is designed as a compromise of the situation that perspective annotations are unavailable, in which the reliability of PENet is essential.
Therefore, we conduct an experiment
 to confirm the feasibility of PENet.
Table~\ref{table:reliablity_of_PENet} demonstrates the comparisons of adopting the ground-truth or the estimated perspective map
as the guidance of spatially variant smoothing on ShanghaiTech Part A/B and WorldExpo'10.
MAEs are respectively 58.1, 9.0 and 8.3, with a small decrease of 1.1, 0.2 and 0.2, respectively.
This indicates that PENet is competent to a reasonable perspective map estimator.

\begin{table}[!h]
	\begin{center}
		\resizebox{1\hsize}{!}{
			\begin{tabular}{ c | c | c  }
				\hline
				Perspective Map & ShanghaiTech Part A/B & WorldExpo'10 \\
				\hline
				Estimated & 58.1/9.0 & 8.3 \\
				\hline
				Ground-truth & \textbf{57.0}/\textbf{8.8} & \textbf{8.1}  \\
				\hline
			\end{tabular}
		}
	\end{center}
	\vspace{-0.5em}
	\caption{Different guidances of PGC on ShanghaiTech Part A/B and WorldExpo'10.}
	\label{table:reliablity_of_PENet}
\end{table}

%%%%%%%%%%%%%%%%%%%%%%%%%%%%%%%%%
%       Sec: More density maps predicted by the proposed PGCNet
%%%%%%%%%%%%%%%%%%%%%%%%%%%%%%%%%
\section{More Density Maps Predicted by the Proposed PGCNet}
Figs.~\ref{fig:more_maps1} and ~\ref{fig:more_maps2} demonstrate more density maps predicted by PGCNet as well as by CSRNet.
From the visualization, it can be seen that our PGCNet shows its superiority to CSRNet in estimating a more accurate number of pedestrians in either sparse or congested scenes.
The quantitative results have been shown in Sec.~5.

%%%%%%%%%%%%%%%%%%%%%%%%%%%%%%%%%%%
%% Fig.D2D
%%%%%%%%%%%%%%%%%%%%%%%%%%%%%%%%%%%
\begin{figure}[!t]
\centering
\includegraphics[width=0.45\textwidth]{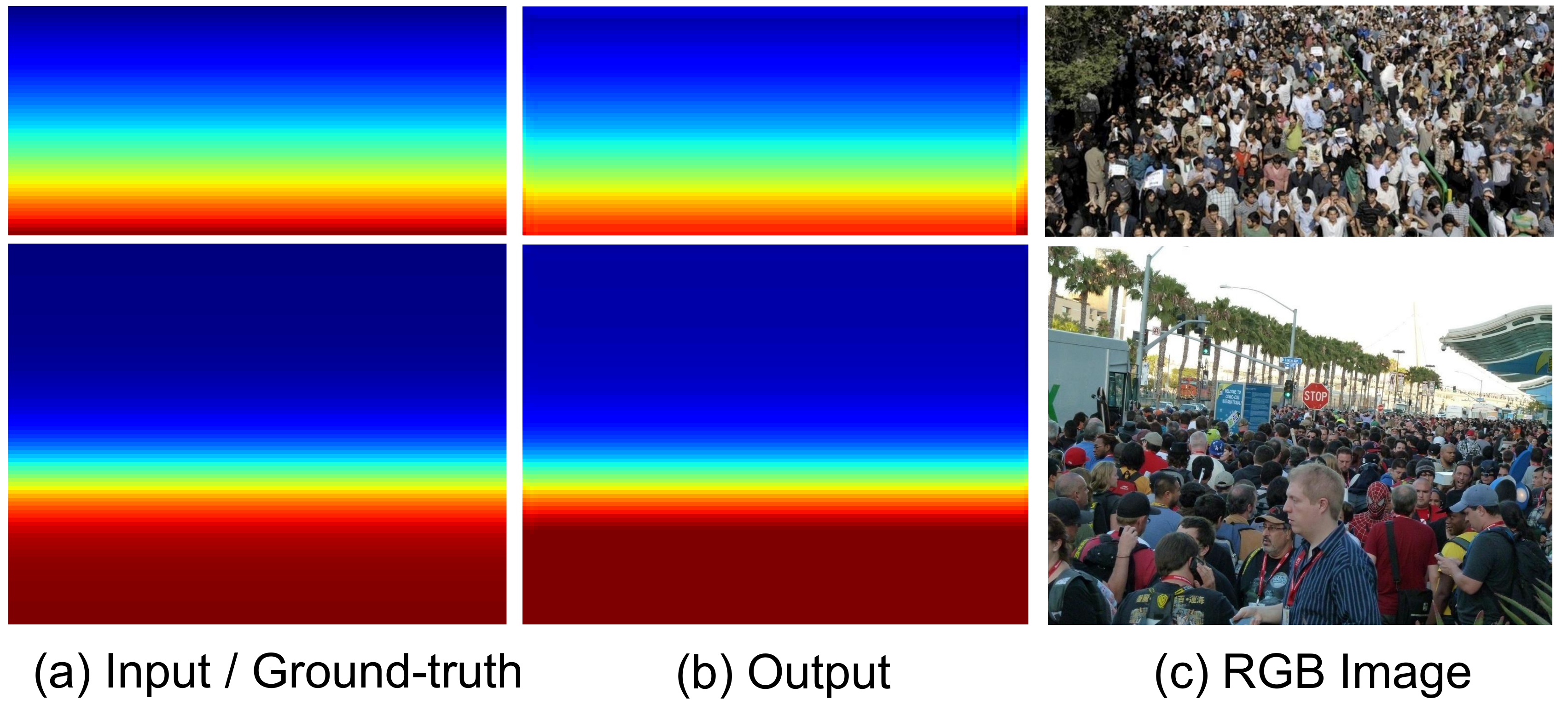}
\caption{Results of the first phase of training PENet. Given (a) the input of PENet, (b) is the reconstructing output, and (c)
              denotes the corresponding RGB image.}
\label{fig:d2d}
\end{figure}

%%%%%%%%%%%%%%%%%%%%%%%%%%%%%%%%%%%
%% Fig.RGB2D
%%%%%%%%%%%%%%%%%%%%%%%%%%%%%%%%%%%
\begin{figure}[!t]
\centering
\includegraphics[width=0.45\textwidth]{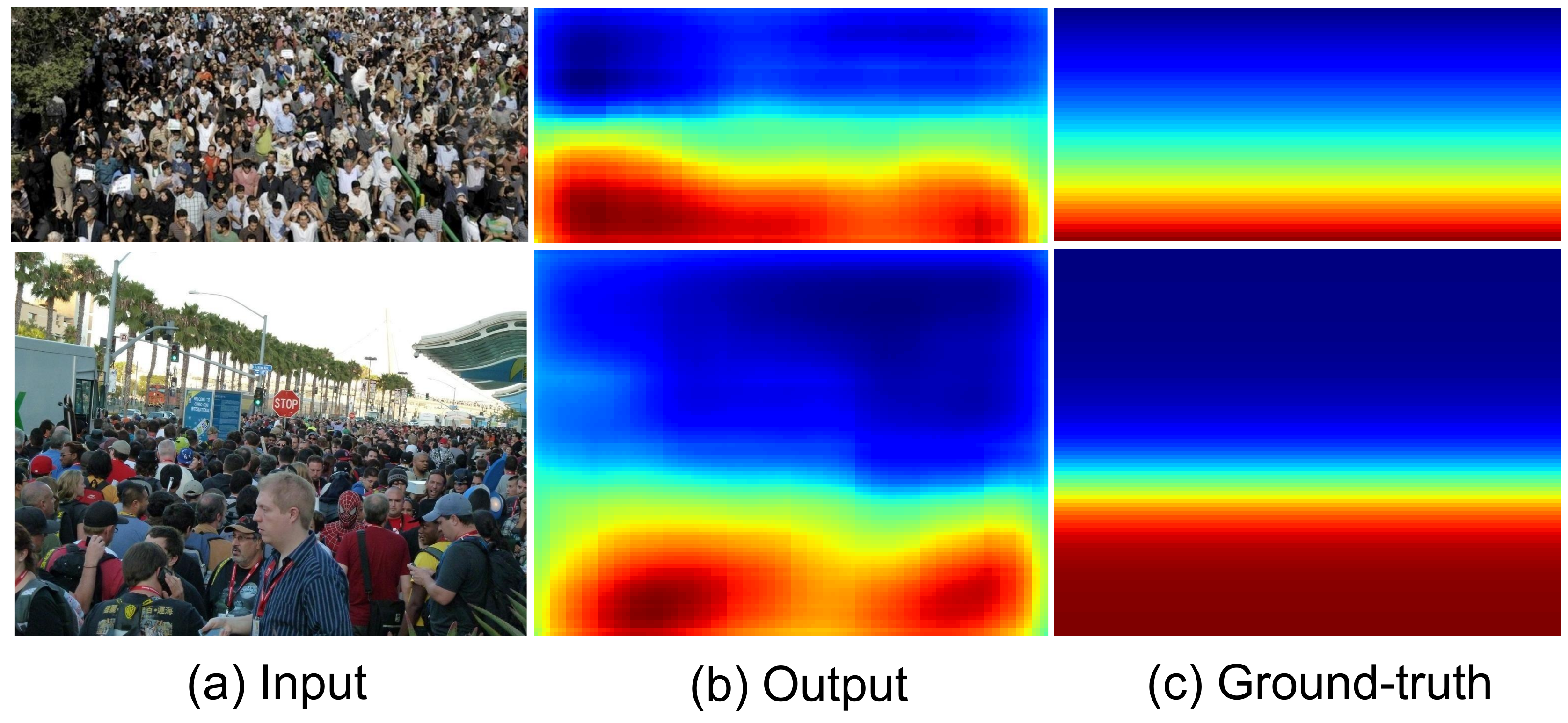}
\caption{Results of second phase of training PENet. Given (a) the input of PENet, (b) is the output of PENet, and (c) represents the ground-truth.}
\label{fig:rgb2d}
\end{figure}

%%%%%%%%%%%%%%%%%%%%%%%%%%%%%%%%%%%
%% Fig.phase3
%%%%%%%%%%%%%%%%%%%%%%%%%%%%%%%%%%%
\begin{figure}[!htb]
\centering
\includegraphics[width=0.45\textwidth]{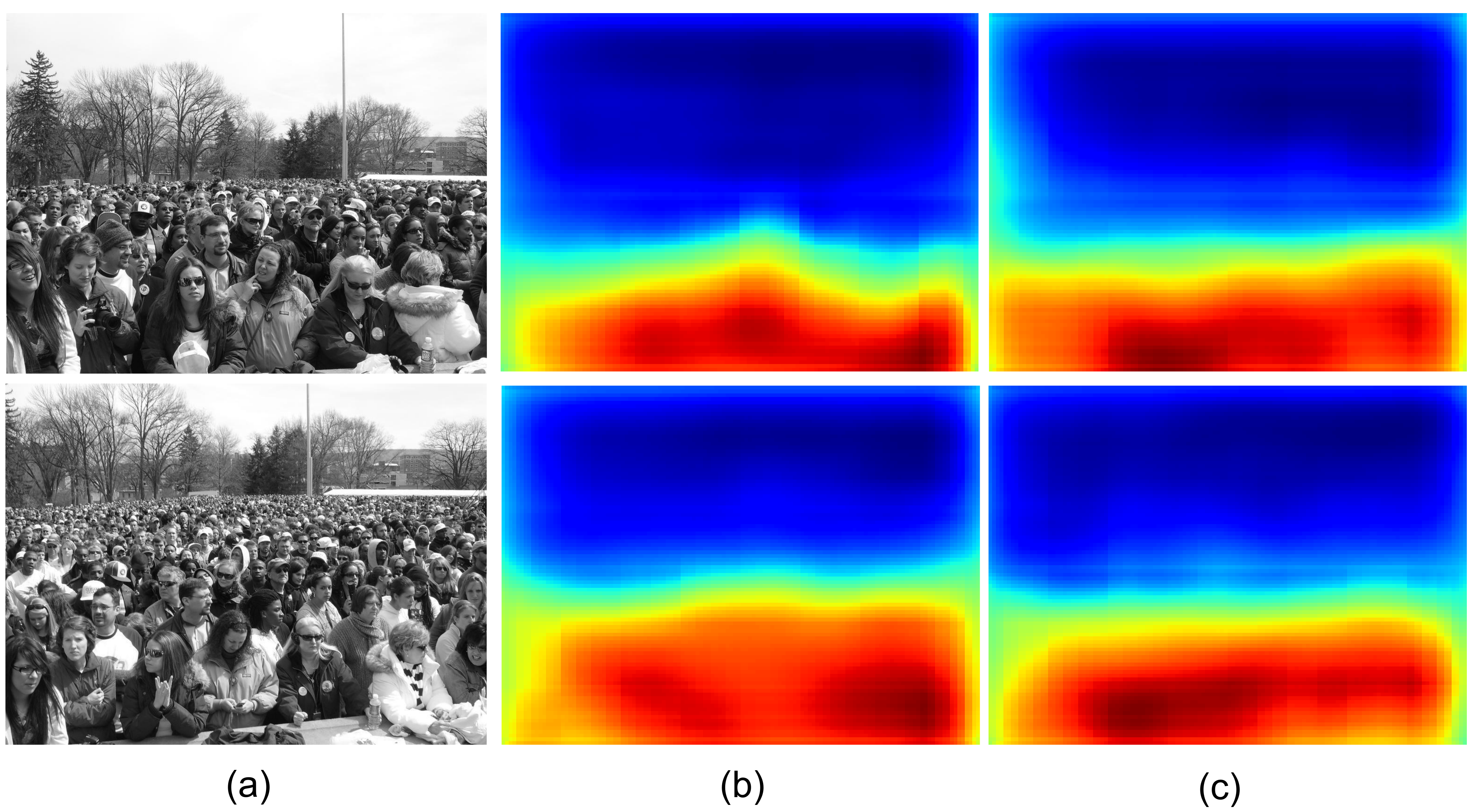}
\caption{Results of third phase of training PENet. Given (a) the original RGB image, (b) is the corresponding perspective map directly predicted by PENet, and (c) represents perspective map when training end-to-end.}
\label{fig:phase3}
\end{figure}

%%%%%%%%%%%%%%%%%%%%%%%%%%%%%%%%%%%%%%%
%%%%   Table: PENet
%%%%%%%%%%%%%%%%%%%%%%%%%%%%%%%%%%%%%%%

\begin{table}[H]
\renewcommand{\arraystretch}{1.00}
\begin{center}
\begin{tabular}{|c|}
\hline
 The architecture of PENet \\
\hline
\hline
Conv. (3, 3, 64), stride=2; LReLU \\
Conv. (3, 3, 128), stride=2;  LReLU   \\
Conv. (3, 3, 256), stride=2; LReLU   \\
Conv. (3, 3, 512), stride=2; LReLU \\
UpConv. (3, 3, 256), stride=2; ReLU   \\
UpConv. (3, 3,128), stride=2; ReLU    \\
UpConv. (3, 3, 64), stride=2; ReLU  \\
UpConv. (3, 3, 1), stride=2;  ReLU  \\
\hline
\end{tabular}
\end{center}
\caption{The architecture of PENet. ``LReLU'' denotes leaky ReLU with the slope of 0.2.}
\label{table:PENet}
\end{table}

%%%%%%%%%%%%%%%%%%%%%%%%%%%%%%%%%%%
%% Fig.More maps 1
%%%%%%%%%%%%%%%%%%%%%%%%%%%%%%%%%%%
\begin{figure*}[t]
\centering
\includegraphics[width=0.9\textwidth]{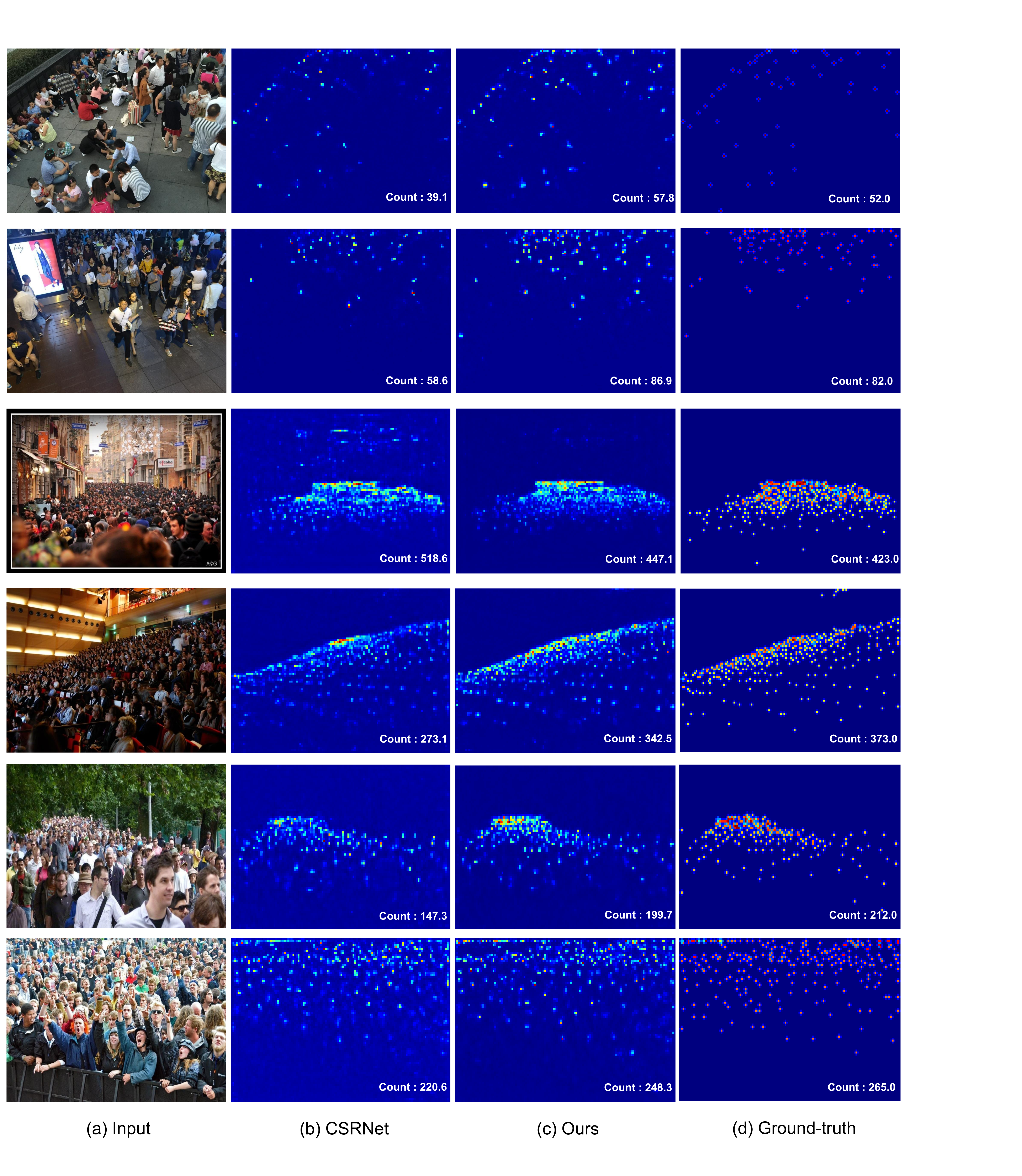}
\caption{Results of density map estimation of CSRNet and our PGCNet.}
\label{fig:more_maps1}
\end{figure*}

%%%%%%%%%%%%%%%%%%%%%%%%%%%%%%%%%%%
%% Fig.More maps 2
%%%%%%%%%%%%%%%%%%%%%%%%%%%%%%%%%%%
\begin{figure*}[t]
\centering
\includegraphics[width=0.9\textwidth]{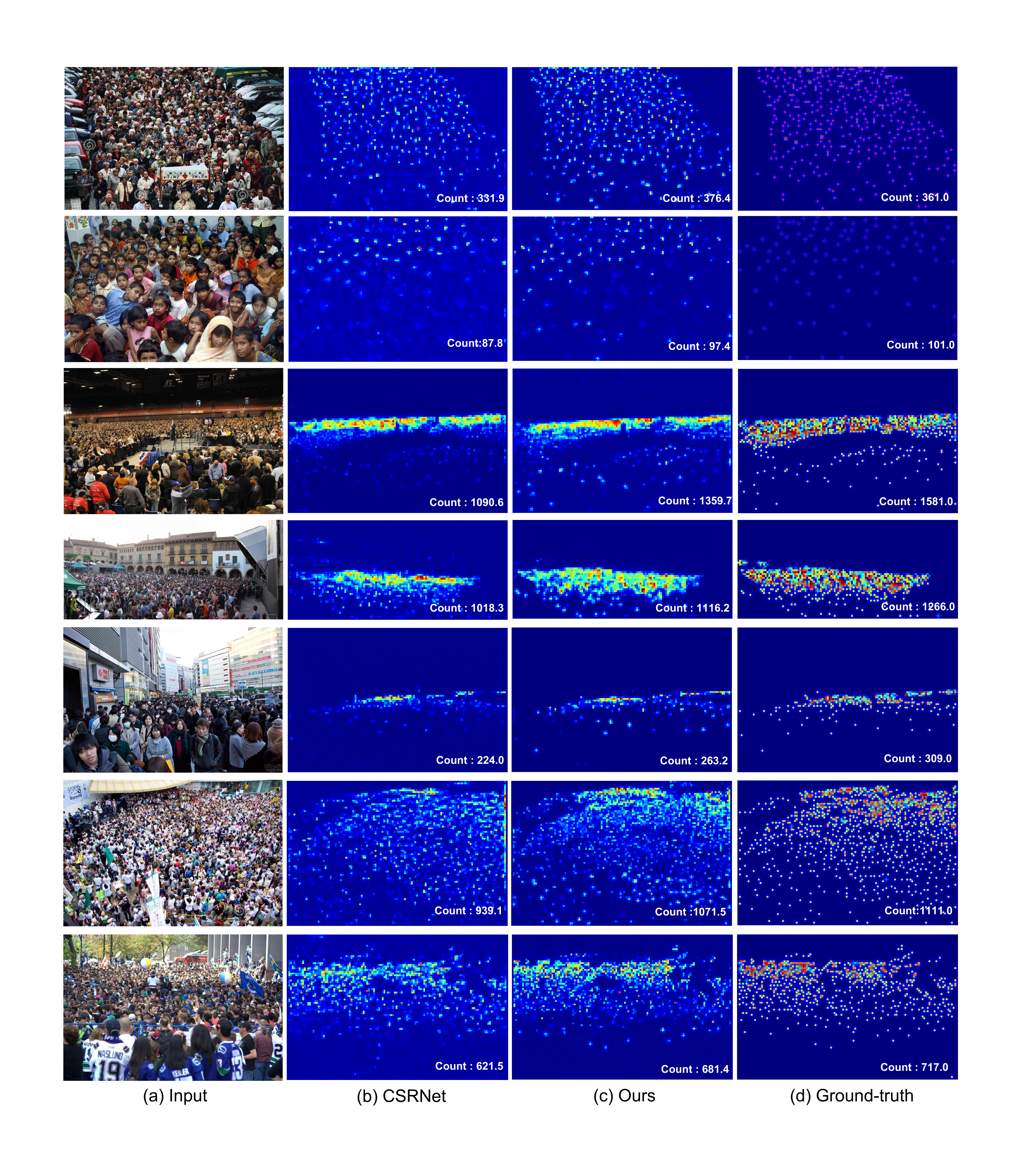}
\caption{Results of density map estimation of CSRNet and our PGCNet.}
\label{fig:more_maps2}
\end{figure*}

\end{appendix}

\end{document}